\documentclass[10pt,letterpaper]{article}

\usepackage[]{wacv}      % To produce the REVIEW version for the algorithms track
\usepackage{graphicx}
\usepackage{amsmath}
\usepackage{amssymb}
\usepackage{booktabs}

\usepackage{caption}
\usepackage{subcaption}
\usepackage{algorithm}
\usepackage{algpseudocode}
\usepackage[pagebackref,breaklinks,colorlinks]{hyperref}

\usepackage[capitalize]{cleveref}
\crefname{section}{Sec.}{Secs.}
\Crefname{section}{Section}{Sections}
\Crefname{table}{Table}{Tables}
\crefname{table}{Tab.}{Tabs.}

\usepackage{times}
\usepackage{epsfig}
\usepackage{graphicx}
\usepackage{caption}
\usepackage{subcaption}
\usepackage{amsmath}
\usepackage{amssymb}
\usepackage{booktabs}
\usepackage{pifont}% http://ctan.org/pkg/pifont
\usepackage{wrapfig}

\def\N{\mathbb{n}}
\def\R{\mathbb{R}}

\def\G{\mathcal{G}}

\def\E{\mathcal{E}}

\usepackage{hyperref}
\usepackage{color}

\def\wacvPaperID{587} % *** Enter the WACV Paper ID here
\def\confName{WACV}
\def\confYear{2025}
\usepackage{times}
\usepackage{graphicx}
\usepackage{caption}
\usepackage{subcaption}
\usepackage{amsmath}
\usepackage{amssymb}
\usepackage{booktabs}
\usepackage{pifont}% http://ctan.org/pkg/pifont
\newcommand\new[1]{{ #1}}

\def\our{GASP}

\def\R{\mathbb{R}}

\def\ga{\mathcal{T}}

\def\G{\mathcal{G}}
\def\N{\mathcal{N}}
\def\m{\mathrm{m}}

\def\G{\mathcal{G}}
\def\N{\mathcal{N}}
\def\m{\mathrm{m}}

\def\V{\mathcal{V}}
\def\E{\mathcal{E}}

\def\m{{\bf m}}
\def\v{{\bf v}}

\def\r{{\bf r}}

\begin{document}

%%%%%%%%% TITLE
\title{\our{}: Gaussian Splatting for Physic-Based Simulations}

\author{Piotr Borycki $^*$\\
Jagiellonian University\\ 
%Jagiellonian University,\\
% Faculty
%of Mathematics and Computer Science,\\
% Cracow, Poland\\
\and
Weronika Smolak $^*$\\
Jagiellonian University\\  
IDEAS\\
\and
Joanna Waczyńska\\
Jagiellonian University\\  
\and
Marcin Mazur \\
Jagiellonian University\\
% Faculty of Mathematics and Computer Science,\\
%  Cracow, Poland\\
\and
Sławomir Tadeja\\
%Department of Engineering, University of Cambridge,\\
Cambridge%, UK.
\and
Przemysław Spurek\\
Jagiellonian University\\
IDEAS Research Institute\\
% Faculty
%of Mathematics and Computer Science,\\
% Cracow, Poland\\
}

\maketitle
\def\thefootnote{*}\footnotetext{These authors contributed equally to this work}\def\thefootnote{\arabic{footnote}}
\thispagestyle{empty}

%%%%%%%%% ABSTRACT
\begin{abstract}
Physics simulation is paramount for modeling and utilizing 3D scenes in various real-world applications. However, integrating with state-of-the-art 3D scene rendering techniques such as Gaussian Splatting (GS) remains challenging. Existing models use additional meshing mechanisms, including triangle or tetrahedron meshing, marching cubes, or cage meshes. Alternatively, we can modify the physics-grounded Newtonian dynamics to align with 3D Gaussian components. Current models take the first-order approximation of a deformation map, which locally approximates the dynamics by linear transformations. In contrast, our GS for Physics-Based Simulations (\our{}) pipeline uses parametrized flat Gaussian distributions.
Consequently, the problem of modeling Gaussian components using the physics engine is reduced to working with 3D points. In our work, we present additional rules for manipulating Gaussians, demonstrating how to adapt the pipeline to incorporate meshes, control Gaussian sizes during simulations, and enhance simulation efficiency. This is achieved through the Gaussian grouping strategy, which implements hierarchical structuring and enables simulations to be performed exclusively on selected Gaussians. The resulting solution can be integrated into any physics engine that can be treated as a black box. As demonstrated in our studies, the proposed pipeline exhibits superior performance on a diverse range of benchmark datasets designed for 3D object rendering.
The project webpage, which includes additional visualizations, can be found at \url{https://waczjoan.github.io/GASP}.
\end{abstract}
%\vspace{-0.7cm}

\begin{figure}[t]
\centering  \includegraphics[width=1.0\textwidth, trim = 10 10 10 170, clip]{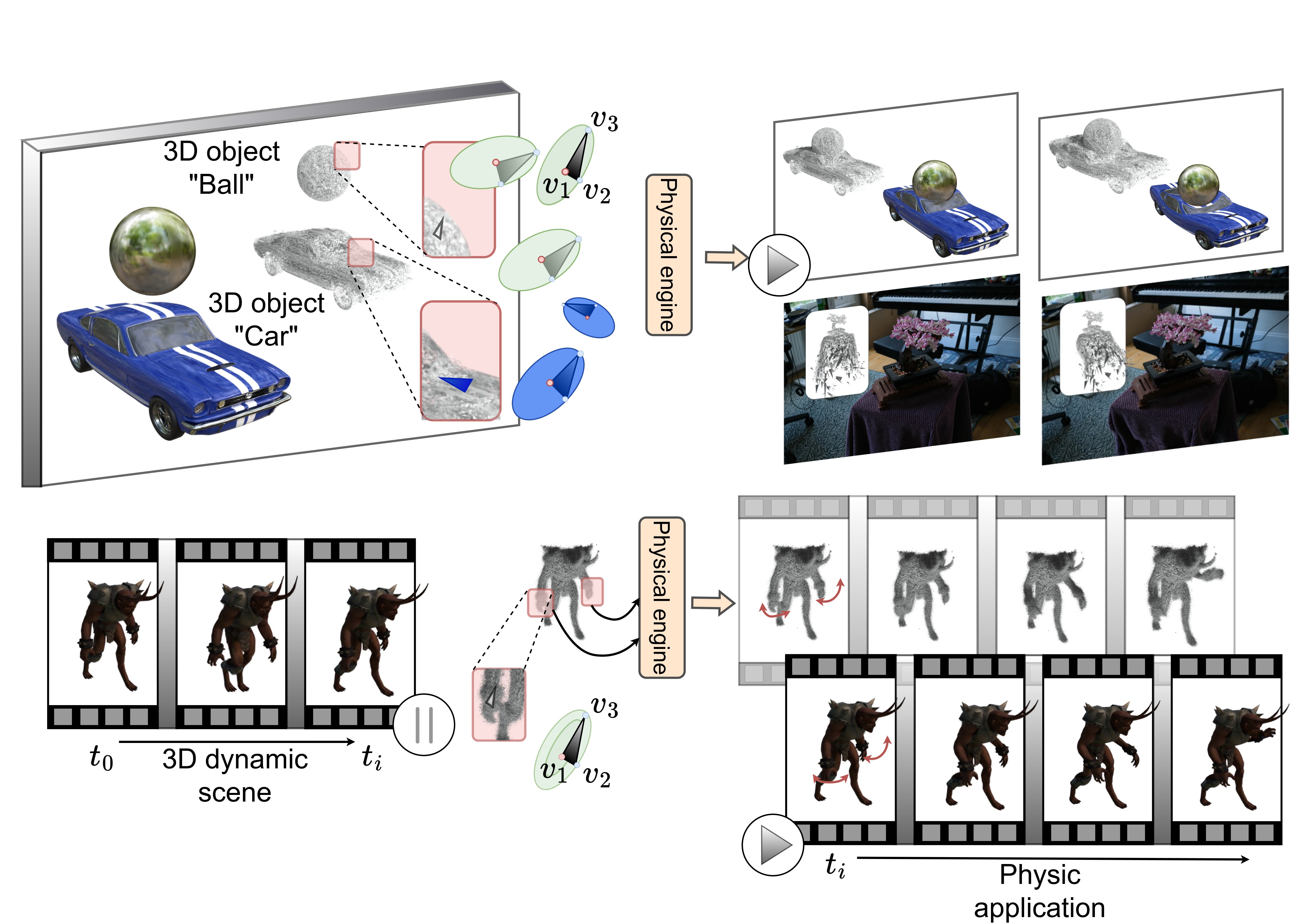}
  \caption{\our{} integrates Gaussian Splatting (GS) with a physics engine to generate realistic simulations. Essentially, \our{} utilizes a set of triangles to parameterize Gaussian components. This approach reduces modifying Gaussian components to 3D point clouds (consisting of vertices of triangles) processing, enabling efficient and rapid rendering. \our{} can model the interaction of objects and work with dynamic scenes.}
  \label{fig:tesser}
\end{figure}

\section{Introduction}

The recent development of \textit{Gaussian Splatting} (GS) \cite{kerbl20233d} has greatly influenced the field of computer graphics, enabling the modeling of three-dimensional (3D) scenes from images with annotated camera positions. The GS model represents 3D scenes as a set of 3D Gaussian distributions, allowing rapid training and rendering. Furthermore, it is relatively straightforward to modify the original GS approach to enable editing~\cite{guedon2024sugar,huang2024gsdeformer,waczynska2024games} and to facilitate the creation of dynamic scenes~\cite{huang2024sc,kratimenos2023dynmf,waczynska2024d}.

%\begin{figure}[ht!]
%    \centering
   % \includegraphics[width=.45\textwidth, trim=0 0 0 0, clip]{img/fox.drawio.png}
    % \caption{\asia{dodać pseudomeshe glowya}}
% \label{fig:fox} 
% \end{figure}

% \begin{figure}[ht!]
%     \centering
%    \includegraphics[width=.46\textwidth, trim=100 50 150 200, clip]{img/samochod.drawio(1).png}
%      \includegraphics[width=.46\textwidth, trim=320 0 0 300, clip]{img/bonsaiobrus.drawio.png}
%        \includegraphics[width=.46\textwidth, trim=100 0 100 0, clip]{img/hellwarrior.drawio.jpg}
%    \caption{\our{} integrates GS with a physics engine to generate realistic simulations. Essentially, \our{} utilizes set of triangle meshes to parameterize Gaussian components. This approach reduces modifying Gaussian components to 3D point clouds (consisting of vertices of triangles) processing, enabling efficient and rapid rendering. \our{} can model the interaction of objects and work with dynamic scenes. 
    % \asia{ja bym gdzies napisala ze kula byla parametryzowana przez mesh prawdziw a nie tylko pseudomesh byl uzywany}
%    }
%\label{fig:tesser} 
%\end{figure}

At the same time, one of the most significant challenges with the GS framework is the incorporation of physics into scenes represented using 3D Gaussian components. 
% Notably, using a physics engine would pave the way to implementing GS in spatial, immersive interfaces such as virtual reality (VR) \cite{jiang2024vr}. 
Existing models employ additional meshing techniques, including triangle or tetrahedron meshing \cite{feng2024gaussian}, marching cubes \cite{guedon2024sugar}, or cage meshes \cite{jiang2024vr}. This approach generates classical mesh-based representations, which are then controlled by a physics engine. Consequently, Gaussian components are adjusted following mesh-based modifications. Although such methods produce satisfactory renderings, they require the implementation of an additional meshing and rendering strategy. \new{Extracting meshes from 3DGS remains a subject of active research, with promising approaches such as SuGaR or 2DGS~\cite{Huang2DGS2024}. However, obtaining smooth, high-quality meshes for arbitrary objects or scenes is still a challenging task, and artifacts affecting the simulation, such as mesh gaps, may occur.}
% Furthermore, modeling tears and cracks represents a challenge, as it is not straightforward to control the behavior of the mesh when the object is divided.

Alternatively, PhysGaussian \cite{xie2024physgaussian} utilizes the vanilla GS approach and adapts Newtonian dynamics to accommodate 3D Gaussian components. PhysGaussian employs the first-order approximation of a deformation map. Using such a method, it can locally approximate the dynamics through linear transformations. This solution is effective, however, it is contingent upon access to a physics engine in order to apply the modifications and relies on a network approximation. Due to that, this method is not suitable for use in other popular 3D graphics software such as Blender.

% \begin{figure}[t]
% \centering
% \includegraphics[width=0.9\columnwidth]{figure1} % Reduce the figure size so that it is slightly narrower than the column. Don't use precise values for figure width.This setup will avoid overfull boxes.
% \caption{Using the trim and clip commands produces fragile layers that can result in disasters (like this one from an actual paper) when the color space is corrected or the PDF combined with others for the final proceedings. Crop your figures properly in a graphics program -- not in LaTeX.}
% \label{fig1}
% \end{figure}

\begin{figure}[t]
    % \centering
    \includegraphics[width=1.0\textwidth, trim=0 200 0 0, clip]{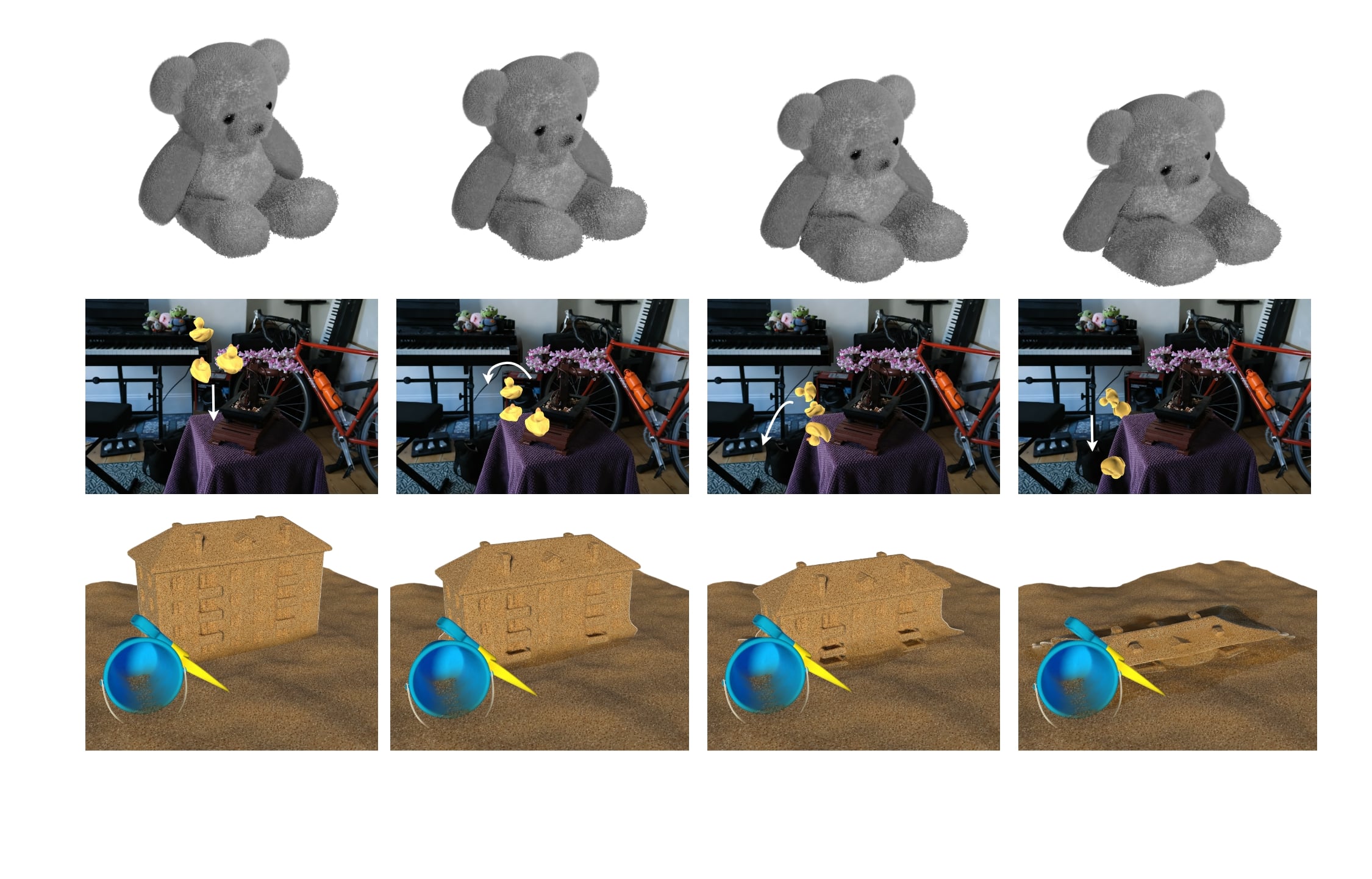}
    \caption{The \our{} pipeline is versatile and compatible with various engines, requiring no significant modifications for integration. We demonstrate its capabilities with simulations performed using the Genesis engine, Blender software, and Taichi Elements.}
    % \Description{The image shows two scenes with interactions between objects. The first row shows the teddy bear falling down. Genesis engine has been used. The second row has the frames from the simulation where three rubber ducks fall on the table and bounce from it. This simulation was made using Blender engine. The third row shows a building made of sand that falls down while the toy sand bucket and shovel stay intact. Taichi\_elements engine has been used for this part. }
\label{fig:kaczki} 
% \vspace{-0.5cm}
\end{figure}

To address these limitations, we propose the {\em Gaussian Splatting for Physics-Based Simulations} (\our{}) pipeline (see Fig.~\ref{fig:tesser}), which incorporates flat Gaussians \cite{waczynska2024d,waczynska2024games} with black box physics engine without special modifications to GS-based model which makes it more flexible compared to other methods.
In this work, we present simulations created using the Genesis\cite{Genesis}, Blender software and the Taichi Elements library (see Fig.~\ref{fig:kaczki}). Additionally, to the best of our knowledge, this is the first demonstration of the integration of a GS-based model with the Genesis engine.
We also demonstrate that straightforward enhancements, such as implementing the Gaussian hierarchy, can seamlessly integrate into our method, significantly accelerating the simulation process.

%Flat Gaussian distributions can be used in GS by applying only minimal changes, such as imposing a zero value on one of the Gaussian components. 
Flat Gaussian distributions can be smoothly employed in GS by modifying Gaussian's scaling component, so it has a zero value on one axis. This representation of GS enables straightforward parametrization and can be easily converted into faces compatible with traditional 3D graphic software. Such techniques have been previously used in several approaches \cite{guedon2024sugar,Huang2DGS2024,waczynska2024games}.
%Such techniques are frequently used in a multitude of approaches \cite{waczynska2024games,guedon2024sugar,Huang2DGS2024}, and they facilitate straightforward modifications to the final model.
The training and rendering times, as well as the quality of the renderings produced by flat Gaussians, are comparable to those of GS.
\our{} approach exemplifies the implementation of Newtonian dynamics within the GaMeS framework, thereby illustrating the capacity to integrate these methodologies without needing external modifications to the physics engine or Gaussian components. More precisely, the model extracts points from GS and applies the Material Point Method to control such points. During the simulation process, the Gaussian components associated with each triple of points are recalculated. This process is notably rapid and does not increase rendering costs. At the same time, treating each point as a discrete entity results in the emergence of artifacts. Consequently, we propose a simple modification that utilizes triangles instead of completely disparate points. We use additional rescaling when a triangle undergoes a significant change in size. In practice, \our{} can produce high-quality simulations in flat GS, including static and dynamic scenes as well as the interaction of objects, see Fig. \ref{fig:tesser}. This is evidenced by the results of experiments conducted on a range of benchmarks with synthetic and real-world data sets.

% \label{fig:bonsai} \label{fig:kaczki} 

%  {\color{red}nie rozumiem skad tu nagle nawaizanie do Gamesa, brak polaczenia z poprzednia czescia paragrfu} Moreover, GaMeS \cite{waczynska2024games} parameterized a flat Gaussian distribution by three points in a 3D space. This allows for the manipulation of the Gaussian component shape through the modification of these points, thereby constituting a method of modifying the GaMeS-based object.

% Furthermore, these points can be maintained, and dynamic scenes can be modeled using a neural network \cite{waczynska2024d}. 
\noindent
The following constitutes a list of our key contributions:

\begin{enumerate}
    % \vspace{-0.1cm}
    \item We propose \our{} pipeline that incorporates physical properties into a 3D scene representation using GS-based models on both static and dynamic scenes.
    % \vspace{-0.2cm}
    \item \our{} operates directly on specific points, thus obviating the necessity for alterations to the physics engine regarded as a black box.
    % \vspace{-0.2cm}
    \item \our{} does not employ additional meshing strategies. We demonstrate that applying additional Gaussian control rules effectively eliminates artifacts, while leveraging hierarchy significantly accelerates simulations.
    %and instead operates directly with parametrized Gaussian distributions.
    
    % . Therefore, we can model the tears and cracks of 3D objects.
    % \vspace{-0.2cm}
\end{enumerate}

\section{Related Work}

The GS representation of 3D scenes is suitable for editing \cite{guedon2024sugar,Huang2DGS2024,waczynska2024games} and modeling dynamic scenes \cite{huang2024sc,wu20234dgaussians,waczynska2024d}. However, the more challenging task is its integration with a physics engine that remains difficult and underexplored. 

\begin{figure*}[t]
    \centering
    \includegraphics[width=1.0\textwidth]{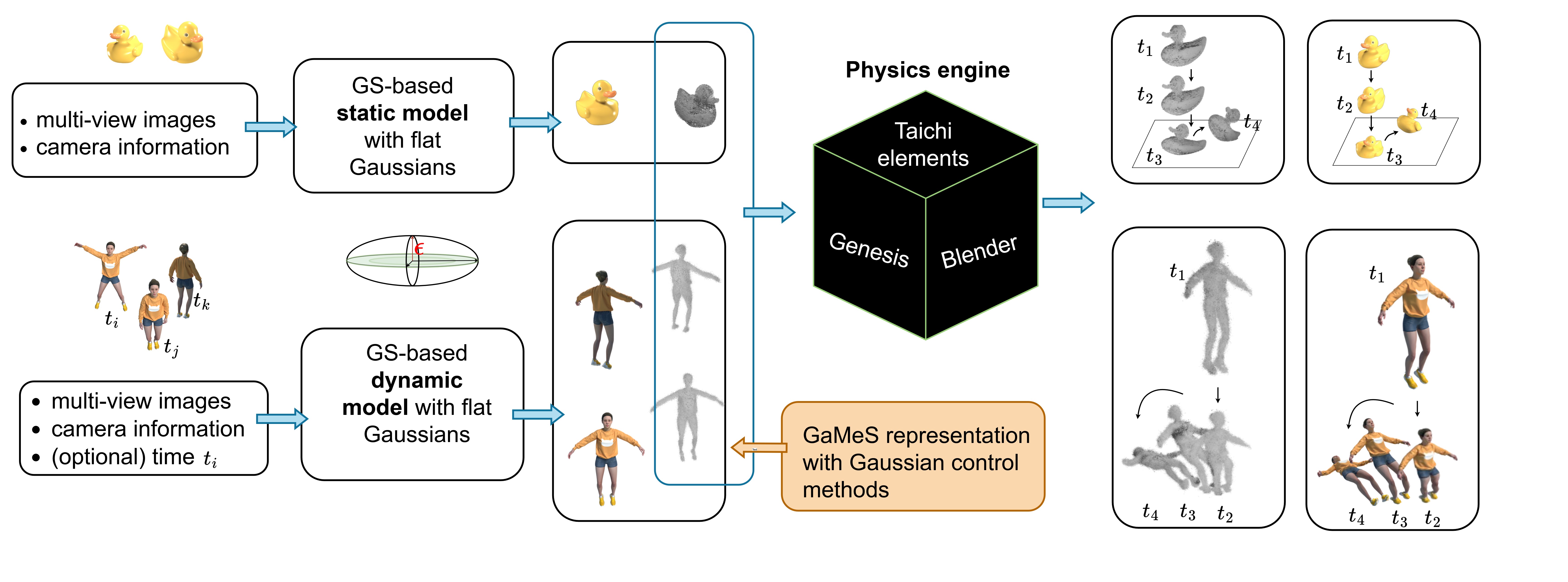}
    % \vspace{-0.5cm}
    \caption{\our{} produces physical simulations using flat GS representation. 
    It works with static GS models like GaMeS~\cite{waczynska2024games} or dynamic ones like D-MiSo~\cite{waczynska2024d}. 
    % In general, \our{} can be applied to any GS-based model which uses flat Gaussians. 
    First, each flat Gaussian component is converted to three points (a triangle face). Then, the physics engine is run on a selected point cloud 
    % \asia{. In our work we show some additional rules which be applied to control Gaussians during simulations}
    to obtain trajectories for our 3D model. Finally, we convert points into Gaussian components by reverse GaMeS parametrization. Notably, we also provided some additional rules that can be applied to control Gaussians during simulations.}
\label{fig:algorithm} 
\end{figure*}

Most existing methods use additional strategic tactics to obtain representations better aligned with physics engines' requirements. Gaussian Splashing (GSP) \cite{feng2024gaussian} is a unified framework that combines 3D Gaussian Splatting
(3DGS) and position-based dynamics.  During the preprocessing, the foreground objects are isolated and reconstructed as a mesh. Such a response is then combined with position-based dynamics. \new{Case-specific methods, such as Embodied Gaussians \cite{pmlr-v270-abou-chakra25a}, introduce a Gaussian Particle that represents an object through particles in order to maintain its geometric aspects for robotics applications.} Alternatively, we can use cage-based representation \cite{jiang2024vr}. \new{Other methods, such as LITA-GS \cite{zhou2025litags}, extract physical attributes from scenes in order to adapt the lighting conditions.}
VR-GS \cite{jiang2024vr} combines 3DGS and eXtended Position-based Dynamics (XPBD) \cite{macklin2016xpbd}. The model starts with scene reconstruction by GS and tetrahedralization. Consequently, we obtain a cage-based representation merged with a physics engine.

All of the above models work with an additional meshing strategy. In contrast, PhysGaussian \cite{xie2024physgaussian} operates directly on Gaussian distributions. The authors of \cite{xie2024physgaussian} suggest approximating a deformation map using the first-order derivative. This yields a linear transformation capable of modifying Gaussian components. Although this model produces relatively good results, it requires adjustments to the physics engine. In contrast, we propose \our{} model, which uses a physics engine as a black box without any modification.

A separate area in physics modeling involves using diffusion models to generate object dynamics. PhysDreamer \cite{zhang2024physdreamer} introduces a method that provides static 3D objects with interactive dynamics by utilizing object dynamics priors acquired from video generation models. Physics3D \cite{liu2024physics3d} learns various physical characteristics of 3D objects using a video diffusion model. DreamPhysics \cite{huang2024dreamphysics} infers the physical attributes of 3DGS based on video diffusion priors.

% \przemek{Ktoś to music napisać}
% video to simulation \cite{liu2024physics3d,huang2024dreamphysics}
% \subsection{Object Representations}
% \cite{kerbl20233d} \cite{waczynska2024games} 

% \subsection{Physic simulation}
% PhysGaussian \cite{xie2024physgaussian} Gaussian Splashing: \cite{feng2024gaussian} NeRFMeshing \cite{rakotosaona2023nerfmeshingdistillingneuralradiance} \cite{modi2024simplicits}

% \subsection{Gaussians Editing}
% \cite{gao2024meshbasedgaussiansplattingrealtime}
% \cite{zhong2024reconstructionsimulationelasticobjects}
% \cite{guedon2024gaussianfrostingeditablecomplex}

\section{\our{} pipeline Overview}

% \begin{figure}[ht]
%     \centering
%     \includegraphics[width=.7\textwidth, trim=0 0 0 0, clip]{img/clamping.drawio.png}
%     \caption{Comparison of physical animation models, illustrating the effects of applying an algorithm for correcting points used in simulation. The left side demonstrates a simulation without the correction algorithm, while the right side shows the enhanced accuracy and stability achieved using Taichi elements with the correction algorithm applied.}
%     % \vspace{-0.5cm}
% \label{fig:artifacts} 
% \end{figure}

% \begin{figure}[ht!]
%     \centering
%     \includegraphics[width=1.0\textwidth, trim=0 0 0 0, clip]{img/foxhier.drawio(1).png}
%     \caption{In the \our{} pipeline, the hierarchical organization of Gaussians through clustering methods allows simulations to focus only on Gaussians at the highest levels of the hierarchy. This approach speeds up computations, especially when using particle-based engines such as Taichi.}
% \label{fig:hier} 
% \end{figure}

\begin{figure}
% \begin{wrapfigure}{R}{0.5\textwidth}
\centering
% \begin{minipage}{.4\textwidth}
  \centering
  \includegraphics[width=0.7\linewidth]{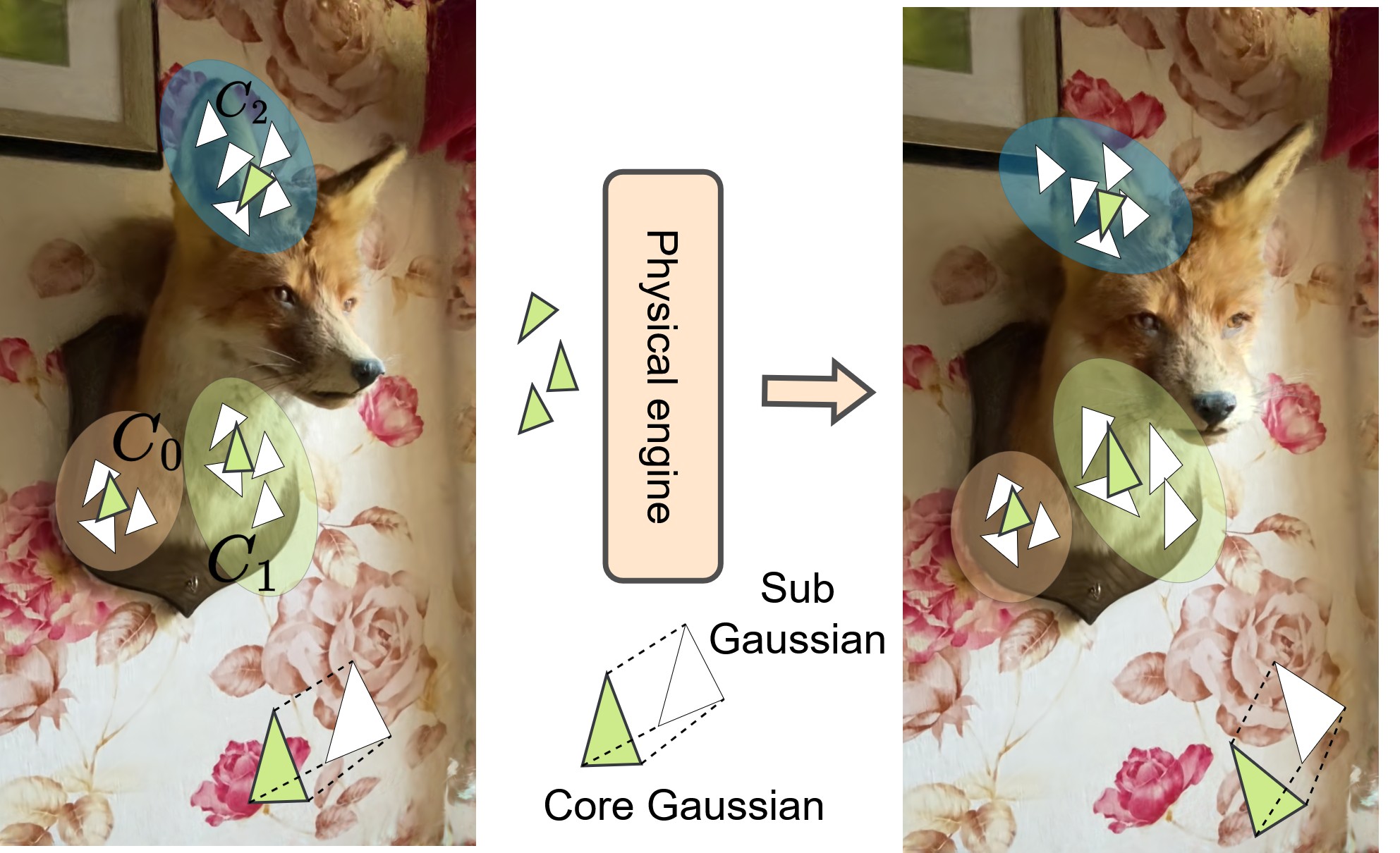}
  \captionof{figure}{In the \our{} pipeline, the hierarchical organization of Gaussians through clustering methods allows simulations to focus only on Gaussians at the highest levels of the hierarchy. This approach speeds up computations, especially when using particle-based engines such as Taichi.}
  \label{fig:hier}
\end{figure}

Here, we present a description of the \our{} pipeline  (see Fig.~\ref{fig:algorithm}). It begins with a general GS model, followed by an examination of the modifications made by GaMeS~\cite{waczynska2024games} to the original GS model, which facilitates straightforward adaptations. Subsequently, the classical MPM~\cite{hu2018moving} is outlined, and finally, our own approach combining both GS and MPM is detailed. In order to improve visual and time quality, we have also proposed two strategies. One is based on the control of the size of Gaussians, and the other is on making appropriate Core Gaussians to create hierarchies (see Fig. \ref{fig:hier}).

\paragraph{GaMeS Parametrization of Gaussian Component}

% \end{minipage}%
    % \qquad
% \end{wrapfigure}

In the classical GS model, each element is characterized by a collection of parameters, including a covariance matrix $\Sigma$ which is factored as
$\Sigma = RSS^TR^T$
where $R$ is the rotation matrix and $S$ is a diagonal matrix containing the scaling parameters. In GaMeS~\cite{waczynska2024games}, on the other hand, flat Guassians are used
$
%\G = \{(
(\N(\m,R,S), \sigma, c ),
$
where $S=\mathrm{diag}(s_1,s_2,s_3)$, with $s_1=\varepsilon$, and $R$ is the rotation matrix defined as
$R=[\r_1,\r_2,\r_3]$, with $\r_i \in \R^3$.
Thus, Gaussian components can be visualized through a triangle-face mesh by utilizing the vertices for their parameterization. We refer to this mapping as $\ga(\cdot)$. In practice, this parameterization results in a set of triangles, which is referred to as triangle soup~\cite{1290060}.
% In {\color{red} our or general?} application, this parameterization results in a set of triangles, termed Triangle Soup~\cite{1290060}.

To outline the GaMeS parameterization, consider a single Gaussian component $\N(\m,R,S)$.
% characterized by the mean $\m$, the rotation matrix $R=[\r_1, \r_2, \r_3]$ and the scaling matrix $S = \mathrm{diag}(\varepsilon, s_2, s_3)$. 
Then its face representation $\N(V)$ is based on a triangle 
$V=[\v_1,\v_2,\v_3]=\ga(\m,R,S)$
with the vertices defined as
% \begin{equation}
$
\v_1 = \m, \; \v_2 = \m + s_2 \r_2, \;  \textrm{and} \;  \v_3 = \m + s_3 \r_3.
$
% \end{equation}
Conversely, given a face (triangle) representation $V=[\v_1,\v_2,\v_3]$, we can recover the Gaussian component:
\begin{equation}
    \N(\hat \m,\hat R,\hat S)=\N(\ga^{-1}(V))
\end{equation}
through the mean $\hat \m$, the rotation matrix $\hat R=[\hat \r_1,\hat \r_2,\hat \r_3]$, and the scaling matrix $\hat S = \mathrm{diag}(\hat s_1,\hat s_2,\hat s_3)$, where the parameters are defined by the following formulas: 
\begin{equation}
    \hat \m = \v_1, \; \; \hat \r_1  =  \frac{(\v_2 - \v_1) \times (\v_3 - \v_1)}{\| (\v_2 - \v_1) \times (\v_3 - \v_1) \|},
\end{equation}
\begin{equation}
    \hat \r_2  =  \frac{(\v_2-\v_1)}{\| (\v_2-\v_1) \|},\;\; \hat \r_3  = \mathrm{orth}(\v_3-\v_1;\r_1,\r_2),
\end{equation}
\begin{equation}
    s_1= \varepsilon, \;\; \hat s_2 = \|\v_2-\v_1\|, \;\; \textrm{and}\;\; \hat s_3 = \langle \v_3-\v_1, \hat \r_3 \rangle.
\end{equation}
Here, $\mathrm{orth}(\cdot)$ denotes a single step of the Gram-Schmidt process~\cite{bjorck1994numerics}.
Accordingly, the corresponding covariance matrix of a Gaussian distribution is given as
$
    \hat \Sigma = \hat R \hat S \hat S^T \hat R^T.
$
% and corresponds to the triangular shape $V$. For a face defined by $(\v_1,\v_2,\v_3)$, the Gaussian component can be expressed as:
% $$ \N((\v_1,\v_2,\v_3)) = \N(\hat \m_V, \hat R_{V}, \hat S_{V}) . $$
% The Gaussian component is derived from the parameters of the mesh face. Consequently, an invertible mapping between Gaussian parameters and the triangular face $T$ will be employed, using the notation:
% $$ 
% \N(V) := \N( \ga^{-1}(V) ) =\N(\hat \m_V, \hat R_{V}, \hat S_{V}). 
% $$

\new{In addition, GaMeS supports distributing Gaussians on meshes when a mesh structure is provided. In this setting, Gaussian means are assigned to each mesh face via learnable barycentric coordinates of its vertices, ensuring they remain bound to the mesh under deformations. The orientation of each Gaussian is defined by the face normal and two in-plane directions derived from the vertices, allowing consistent updates when the mesh position changes. In this work, we focus on the mesh-free case, though the proposed formulations are directly applicable to Gaussians placed on mesh faces.}

\begin{figure}[t]
    \centering
    
    \centering
  %\begin{minipage}[t]{0.45\textwidth}
    \begin{tabular}[h]{ccccc}
  &  $t_0$ & $t_i$ & $t_j$ & $t_T$ \\
   \raisebox{3.5\normalbaselineskip}[0pt][0pt]{\rotatebox[origin=c]{90}{2DGS}}  &
    \includegraphics[width=0.2\linewidth, trim=50 50 50 200, clip]{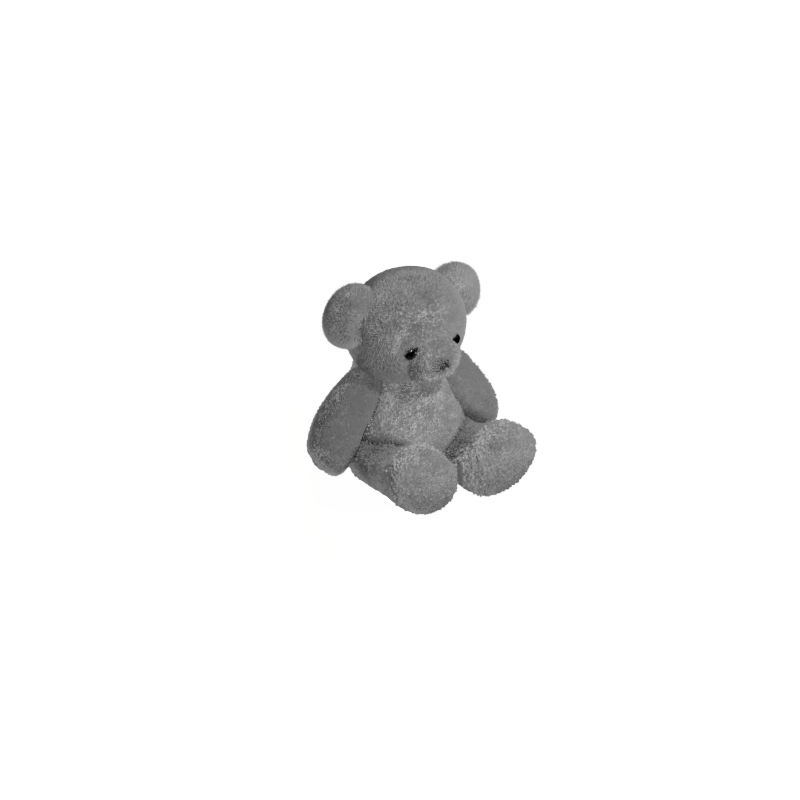} &
    \includegraphics[width=0.2\linewidth, trim=50 50 50 200, clip]{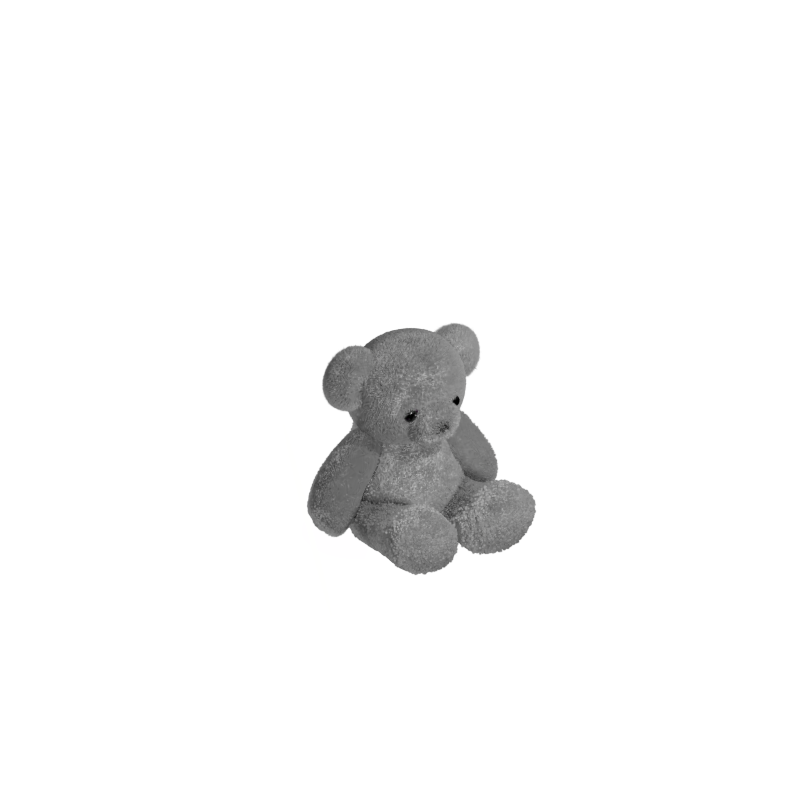} &
    \includegraphics[width=0.2\linewidth, trim=50 50 50 200, clip]{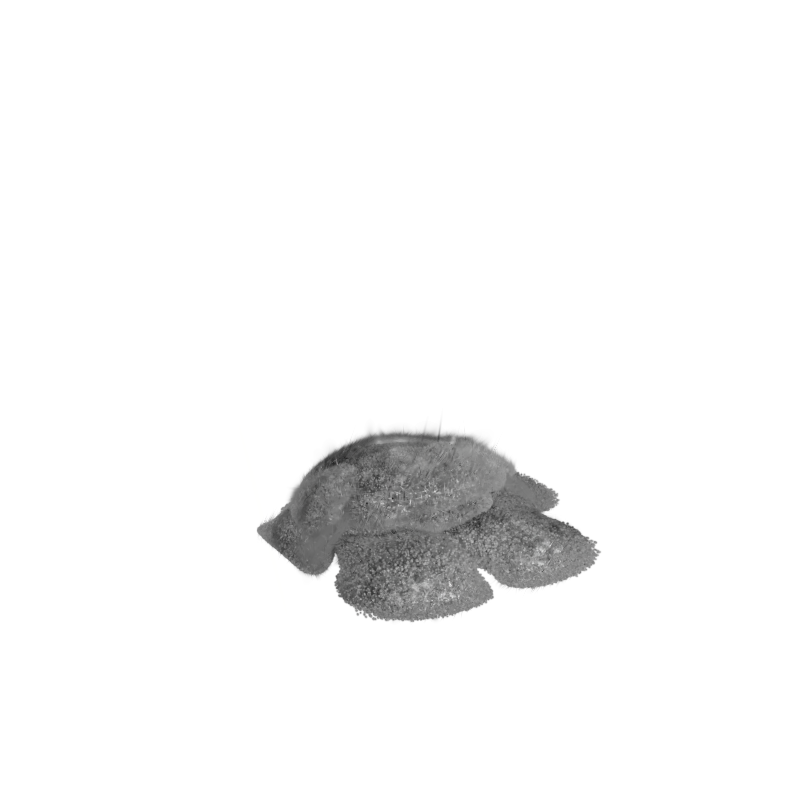} &
    \includegraphics[width=0.2\linewidth, trim=50 50 50 200, clip]{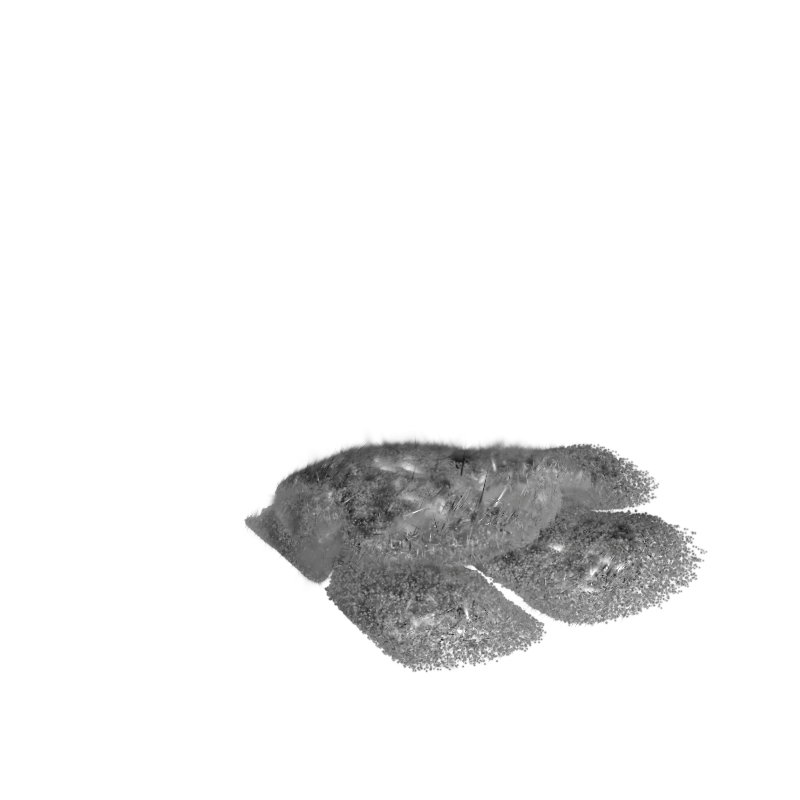}
\\ \raisebox{3.5\normalbaselineskip}[0pt][0pt]{\rotatebox[origin=c]{90}{GaMeS}}  &
    \includegraphics[width=0.2\linewidth, trim=50 50 50 200, clip]{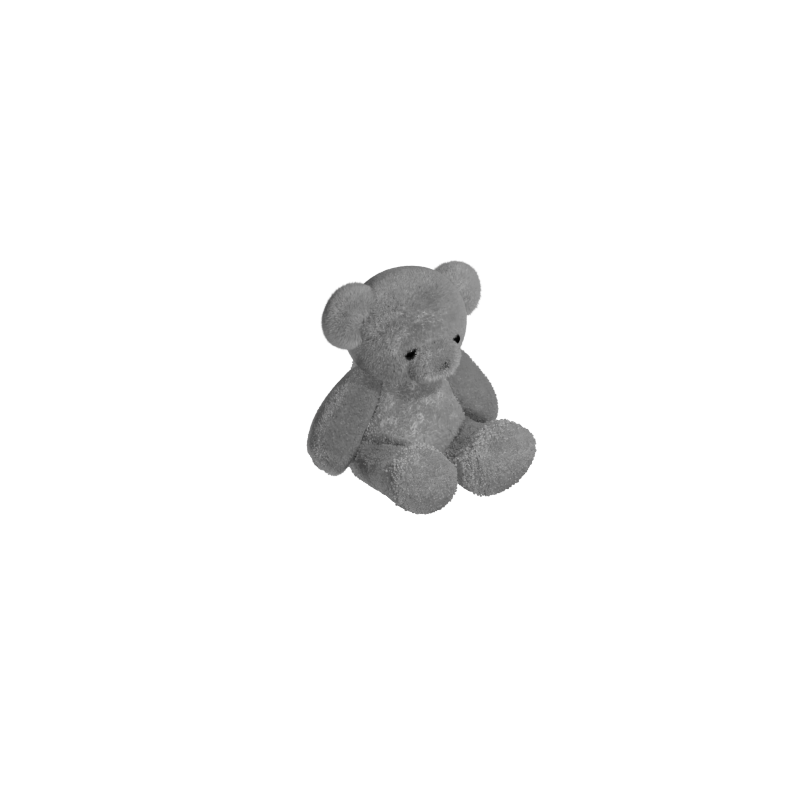} &
    \includegraphics[width=0.2\linewidth, trim=50 50 50 200, clip]{img/rebuttal/our/03_0000.png} &
    \includegraphics[width=0.2\linewidth, trim=50 50 50 200, clip]{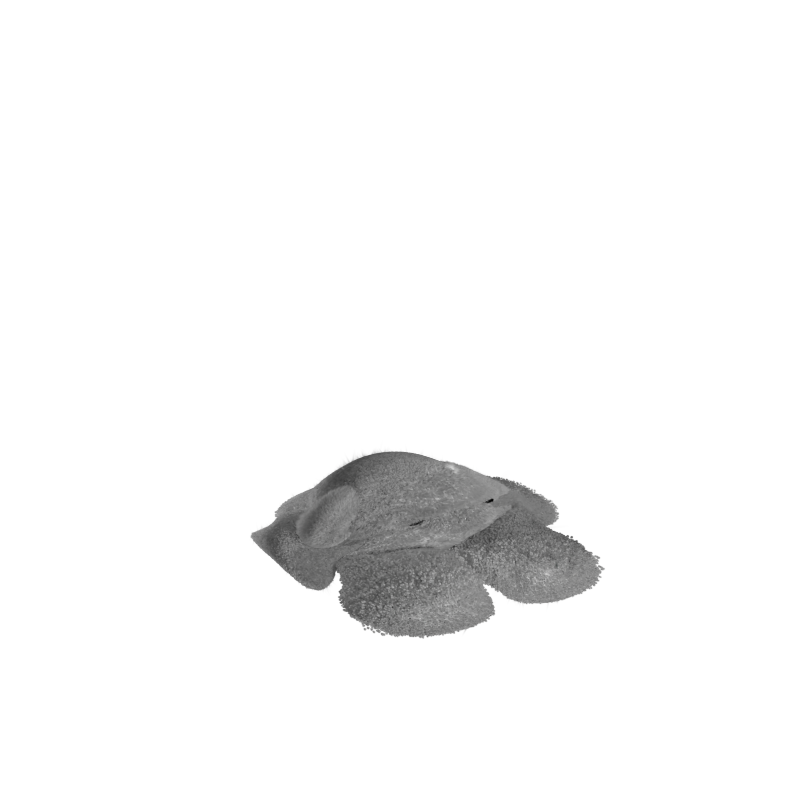}&
    \includegraphics[width=0.2\linewidth, trim=50 50 50 200, clip]{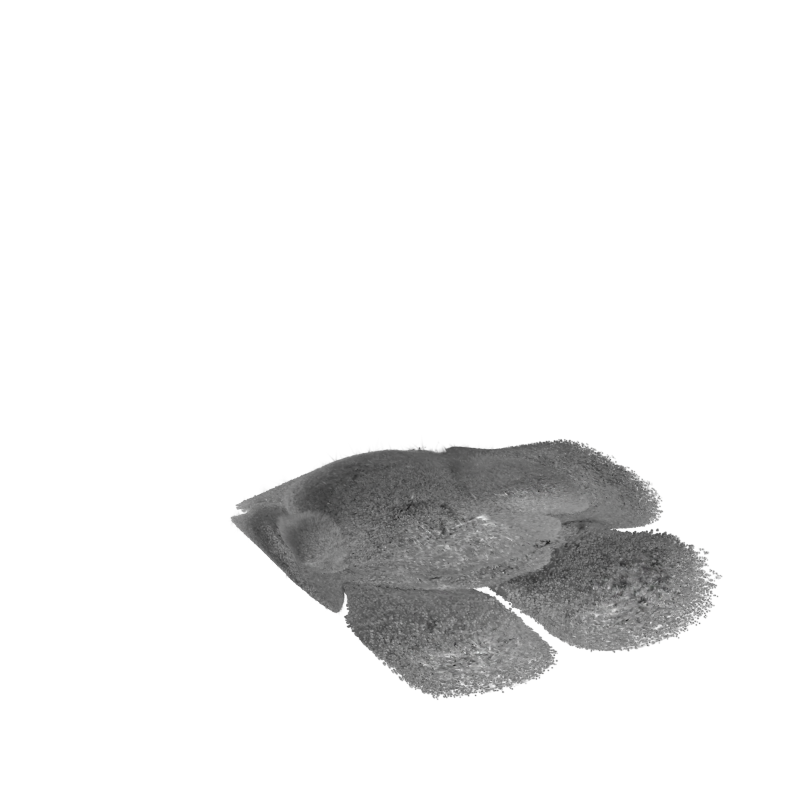}
    
    \end{tabular}

    \caption{
    \new{Visual comparison of gravity simulations with sand-like material represented by points. First row: direct application of physical simulators to Gaussians obtained from 2DGS, which produces noticeable artifacts. Second row: application of physical simulators on triangle faces obtained using the GaMeS parametrization, the pointed artifacts and holes are not as apparent.}
    }
    \label{fig:comp_2dgs}
\end{figure}

\paragraph{Constitutive Model}

% In alignment with the methodology delineated by \cite{xie2024physgaussian}, our constitutive model is founded upon the tenets of continuum mechanics. 
In alignment with the methodology delineated by \cite{xie2024physgaussian}, the tenets of continuum mechanics form the foundation of the constitutive model.
The fundamental principles of continuum mechanics entail the representation of material deformation through the use of a map $\phi$, which transforms points from the initial undeformed material space $X$ to the subsequently deformed world space $x=\phi(X,t)$ at time $t$. This process yields the following deformation gradient:
\begin{equation}\label{eq:defgradient}
    F(X, t) = \nabla_X\phi(X, t),
\end{equation}, which is used to measure local rotation and strain. The deformation function $\phi$ is subject to changes in accordance with the principles of conservation of mass, conservation of momentum, and the elasto-plastic constitutive relation: 

\begin{equation}\label{eq:conservation}
\frac{D\rho}{Dt}=0,\;\; \frac{Dv}{Dt}=\frac{1}{\rho}\nabla\cdot \sigma +f,
\end{equation}
% \begin{equation}\label{eq:momentumconservation}
% \frac{Dv}{Dt}=\frac{1}{\rho}\nabla\cdot \sigma +f,    
% \end{equation}
and \begin{equation}\label{eq:elasto}
\sigma = \frac{1}{\det(F)} \frac{\partial \Psi }{\partial F^E} ({F^E})^T.    
\end{equation}

In this context, $\rho$ and $v$ denote the density and velocity as functions of time $t$ and position $x$, while $\sigma$ is the Cauchy stress tensor, which includes both pressure and shear, and \mbox{$\nabla \cdot \sigma$} denotes its divergence. Furthermore, $f$ represents a vector that includes all accelerations caused by body forces (e.g. gravitational acceleration). On the other hand, $\Psi$ is the strain energy density function of $F$, which quantifies the amount of non-rigid deformation. Note that in our work, we use a plasticity model in which the deformation gradient is multiplicatively decomposed (following some yield stress condition) into an elastic part $F^E$ and a plastic part $F^P$, so that $F = F^EF^P$.

\paragraph{Material Point Method}
The fundamental premise of the MPM is the utilization of particles, or material points, for the tracking of mass, momentum, and deformation gradient (here, we track rather $F^E$ than $F$). From one perspective, the Lagrangian approach enables the monitoring of time-varying characteristics, such as position, velocity, and deformation gradient, in alignment with the mass conservation principle for each particle. This approach ensures the constancy of the overall mass. Conversely, the lack of mesh connectivity between particles represents a substantial challenge in calculating derivatives, which are crucial for stress-based force evaluation (see Equation~\eqref{eq:elasto}) and are, in turn, a fundamental component of the momentum conservation principle. This issue can be addressed by employing a regular Eulerian grid as a background and discretizing the Cauchy stress tensor divergence $\nabla\cdot \sigma$ with the use of interpolation functions over the grid in a manner analogous to the standard Finite Element Method (FEM), utilizing the weak form. In this approach, the grid basis functions are dyadic products of one-dimensional cubic B-splines \cite{steffen2008analysis}.

The MPM algorithm is a three-stage process comprising particle-to-grid transfer, grid update, and grid-to-particle transfer, which are applied in a loop \cite{xie2024physgaussian}. In the initial stage, mass and momentum are transferred from the particle to the grid. In the subsequent stage, grid velocities are updated based on forces at the next timestep. In the final stage, the updated grid velocities are transferred back to the particles, and then new quantities of the particles are computed.

\begin{figure}[t]
    \centering
    \includegraphics[width=0.9\textwidth, trim=0 0 0 0, clip]{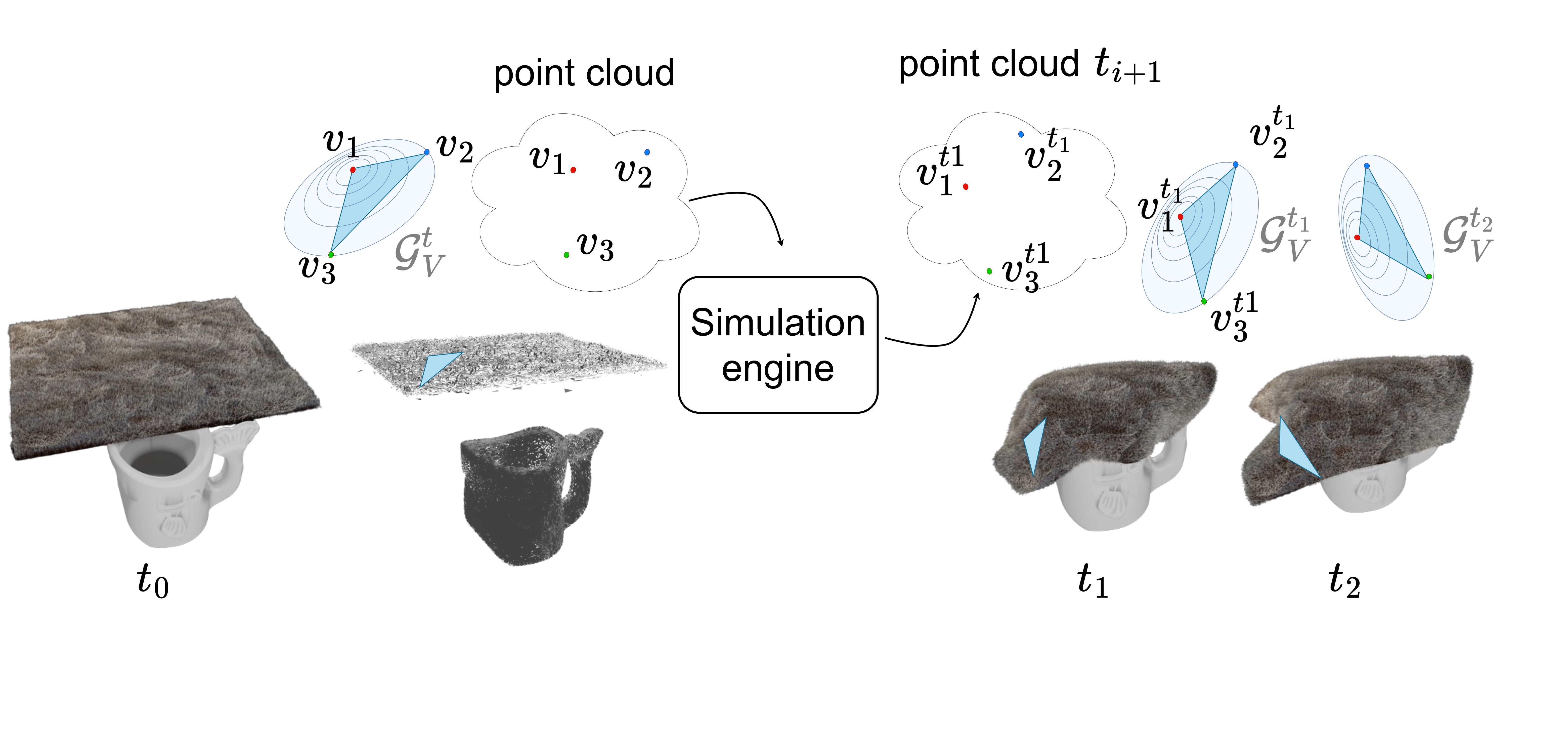}
    % \vspace{-0.8cm}
    \caption{ \new{\our{} operates directly on the Gaussian triangle representation without altering the physics-based simulation engine.} Initially, the 3D objects are depicted as flat Gaussians. GaMeS then converts these Gaussians into triplets (mesh faces). The simulation engine transforms these points following the MPM model. Subsequently, the new points are reassembled into triangles and then into Gaussians. This process is speedy, yielding realistic renderings.}
\label{fig:gaussmoves} 
\end{figure}

\paragraph{\our{} -- Physics in GaMeS}

This subsection presents a description of the \our{} pipeline. Since the model employs the GaMeS representation of a 3D scene, we postulate that the input comprises a collection of Gaussians:
\begin{equation}
\G = \{(\N(\m_i,R_i,S_i), \sigma_i, c_i )\}_{i=1}^{p}, 
\end{equation}
with means $\m_i$, scaling matrices $S_i$, and rotation matrices $R_i$. 
%$S=\mathrm{diag}(s_1,s_2,s_3)$, $s_1=\varepsilon$,
%and $R$ is the rotation matrix.
Furthermore, we assume the existence of a deformation map $\phi(X,t)$, which models the behavior of 3D objects. To obtain $\phi(X,t)$, it is necessary to have a point cloud that describes 3D objects. As our method employs the GaMeS reparametrization of GS, we can transform the mean Gaussian parameters and the covariance matrix into a mesh (triangle soup). In such an instance, we may consider the system $(\V, \E)$, where:
\begin{equation}
\V = \{ V_i= \ga(\m_i,R_i,S_i)  \}_{i=1}^{p}
\end{equation}
denotes a collection of vertices, and $\E$ represents the edges obtained via GaMeS. \new{Our physics-based simulation engine operates at each point of $V_i$ individually.} Subsequently, the vertices of the triangle soup can be used as input for the MPM algorithm in conjunction with tools such as the Blender software, Taichi\_elements, or Genesis libraries (for further details, see the experimental section), see Fig.~\ref{fig:kaczki}.
It is essential to note that our solution does not entail any modifications to the MPM, in contrast to the Taylor approximation employed in \cite{xie2024physgaussian}.
Consequently, in each time step, the following is obtained:
\begin{equation}
    \V_t = \{ \phi(V_i,t) \colon  V_i \in \V \}.
\end{equation}
The triangle soup in time $t$ is then defined by the collection of vertices and edges $(\V_t,\E_t)$. In contrast, the reverse parametrization provided by GaMeS can be employed to obtain Gaussian Splatting:
\begin{equation}
\G_t = \{( \N( \ga^{-1}( \phi(V_i,t) )), \sigma_i, c_i )\}_{i=1}^{p}. 
\end{equation}

\new{Our \our{} pipeline represents Gaussian components through triangle faces and integrates them with a point-based physics engine. Flat Gaussian components are first converted into triangle faces, and the simulation is then performed on their vertices. Direct application of standard physics simulators often produces noticeable artifacts (see Fig.~\ref{fig:comp_2dgs}).  In contrast, our approach effectively supports the learned flat Gaussians by transforming them into triangles, ensuring stable simulations. Unlike alternative flat Gaussian representations, such as 2DGS, the GaMeS parametrization naturally fits this setting, since each flat Gaussian is inherently represented by a triangle.
}

\begin{figure}[t]
%\begin{wrapfigure}{R}{0.5\textwidth}
% \begin{minipage}{.4\textwidth}
  \centering
  \includegraphics[width=0.59\linewidth]{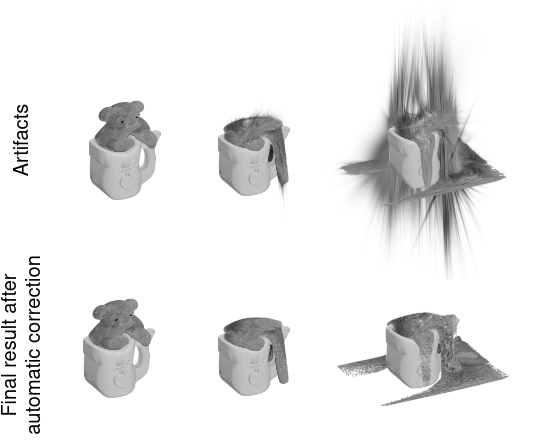}
  \captionof{figure}{Comparison of physical animation models, illustrating the effects of applying an algorithm to correct points used in simulation. The left side demonstrates a simulation without the correction algorithm, while the right side shows the enhanced accuracy and stability achieved using Taichi elements with the correction algorithm applied.}
  \label{fig:artifacts}
% \end{minipage}
\end{figure}
%\end{wrapfigure}

\begin{figure}[b]
    \centering
    \includegraphics[width=0.9\textwidth, trim=0 0 0 0, clip]{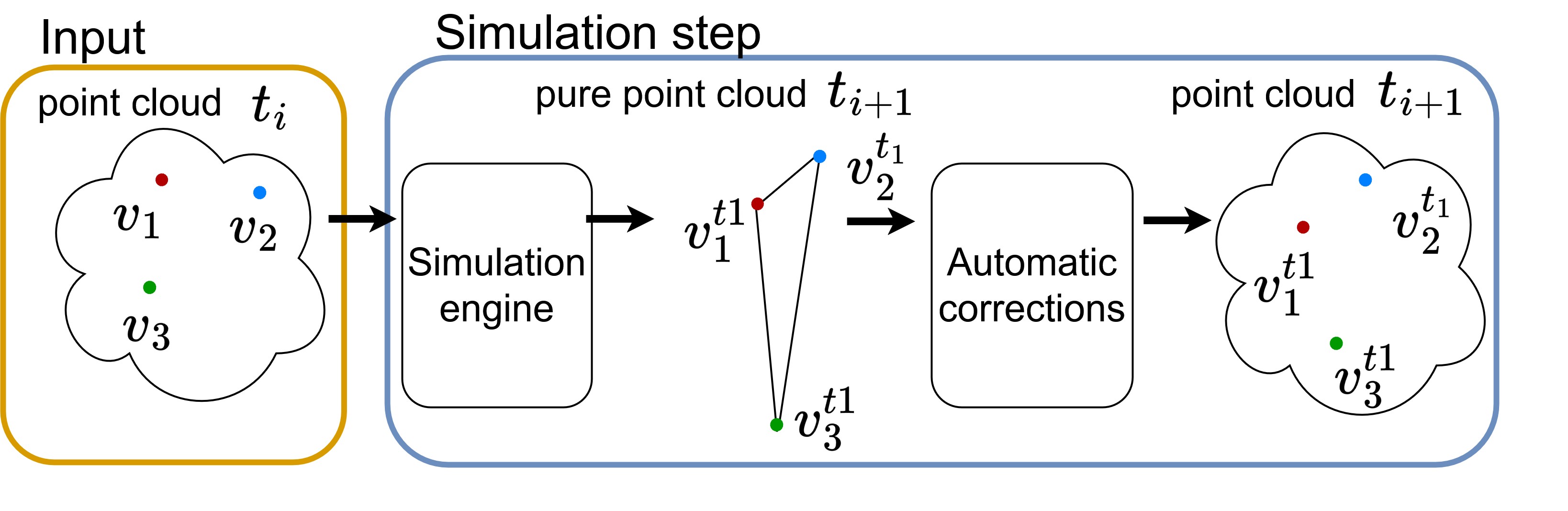}
    \caption{\our{} works directly on points cloud extracted from flat Gaussians. In practice, physics engine like Taichi\_elements moves all particles independently, which can produce artifacts. Therefore, we use a correction algorithm to control the distance of points from each other. If some points are moved extremely far, we automatically correct such behaviors by clipping parameters $S$ in GS. }
\label{fig:simulationstep} 
% \vspace{-0.5cm}
\end{figure}

\paragraph{Gaussian Control Rules}
It is important to note that processing each vertex of a triangle soup individually is a highly rapid process that treats the physics engine as a black box. The schema of such processes is presented in Fig.~\ref{fig:gaussmoves}. Our method is compatible with any physics engine that operates on three-dimensional points. However, the processing of all points separately may yield artifacts. When all points are treated independently during simulation, some elements exhibit considerable fluctuations in velocity. This phenomenon leads to the formation of extensive triangles (mesh faces), which in turn give rise to prominent Gaussian components, as illustrated in Fig.~\ref{fig:artifacts}. To mitigate this effect, our approach employs a straightforward yet highly effective technique (correction algorithm). Specifically, we monitor the distance between 
% points that correspond to the principal components of the normal distribution from the center. 
vertices of the face representation $V=[\v_1,\v_2,\v_3]$ of a Gaussian component $\N(\m,R,S)$.
It should be noticed that, in the case of GaMeS transformation, the first vertex
% , designated {\color{red}designated vertex? obecnie troche to nie ma sensu} 
is equal to the center of the distribution ($\v_1=\m$), while the two subsequent vertices are located on the main components of the ellipses that describe the distribution ($\v_2 = \m + s_2 \r_2$, $\v_3 = \m + s_3 \r_3$). Consequently, if the distance between the transformed mesh nodes is considerably greater than that of the original nodes, i.e.:
\begin{equation}
\label{eq:alpha_rule}
\|  \phi(\v_1,t) - \phi(\v_i,t) \| > \alpha \|  \v_1 - \v_i \| \;\;\mbox{for}\;\; i=2,3,
\end{equation}
where $\alpha$ is a hyperparameter, we automatically assert that the value of $s_i$ should be identical to that of the original value times $\alpha$. This is illustrated in Fig.~\ref{fig:simulationstep}.
% In all experiments, we set the $\alpha$ to 1.5. \piotrek{To zdanie chyba nie jest prawda} 
% It should be noted that throughout the full simulation, both the mesh and GS are accessible, enabling real-time rendering.

\paragraph{Gaussians Hierarchy in Simulations}
%%%%%%%%%%%%%%%%%%%%%%%%%%%%5
%%%%%%%%%%%%%%%%%%%%%%%%
%%%%%%%%%%%%%%%%%%%%%%%%%%%%%%
%%%%%%%%%%%%%%%%%%%%%%%%%%%%%%
%%%%%%%%%%%%%%%%%%%%%%%%%%%%%%%%

% \begin{figure}[ht!]
\begin{wrapfigure}{R}{0.5\textwidth}
    \centering
    \includegraphics[width=0.45\textwidth, trim=0 0 0 0, clip]{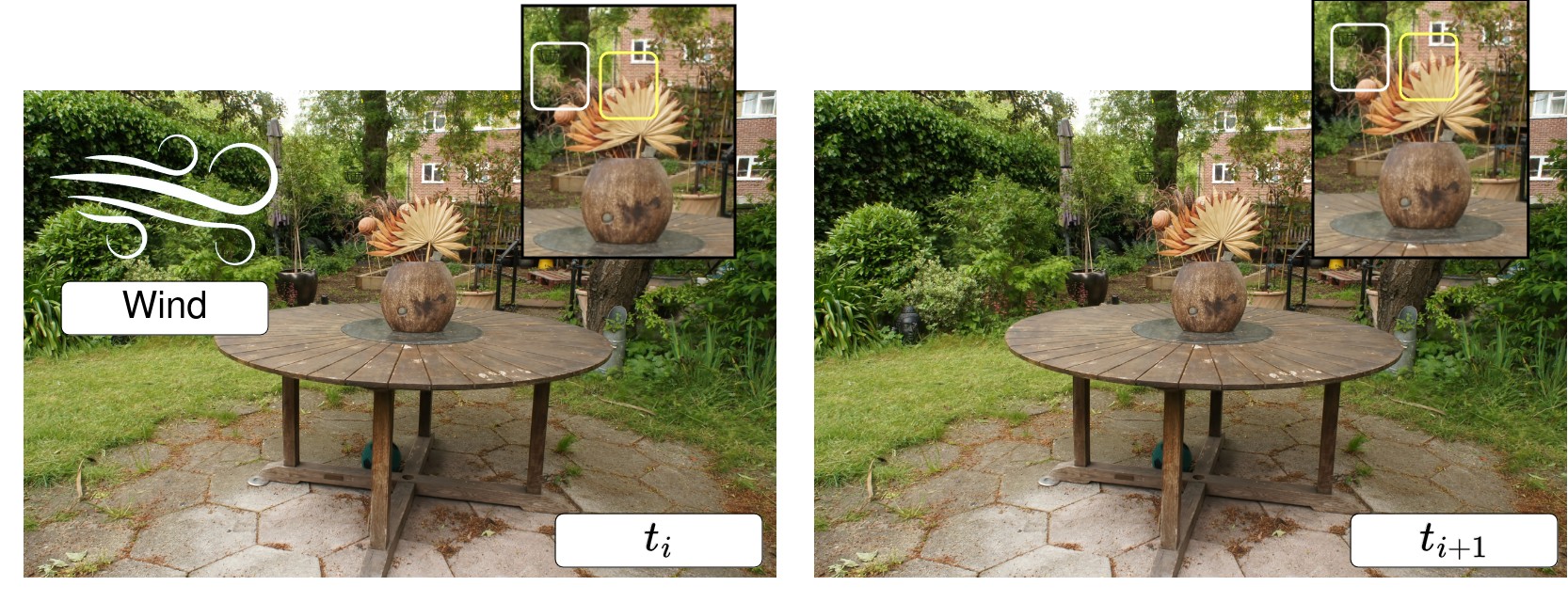}
    \includegraphics[width=.45\textwidth, trim=0 0 0 0, clip]{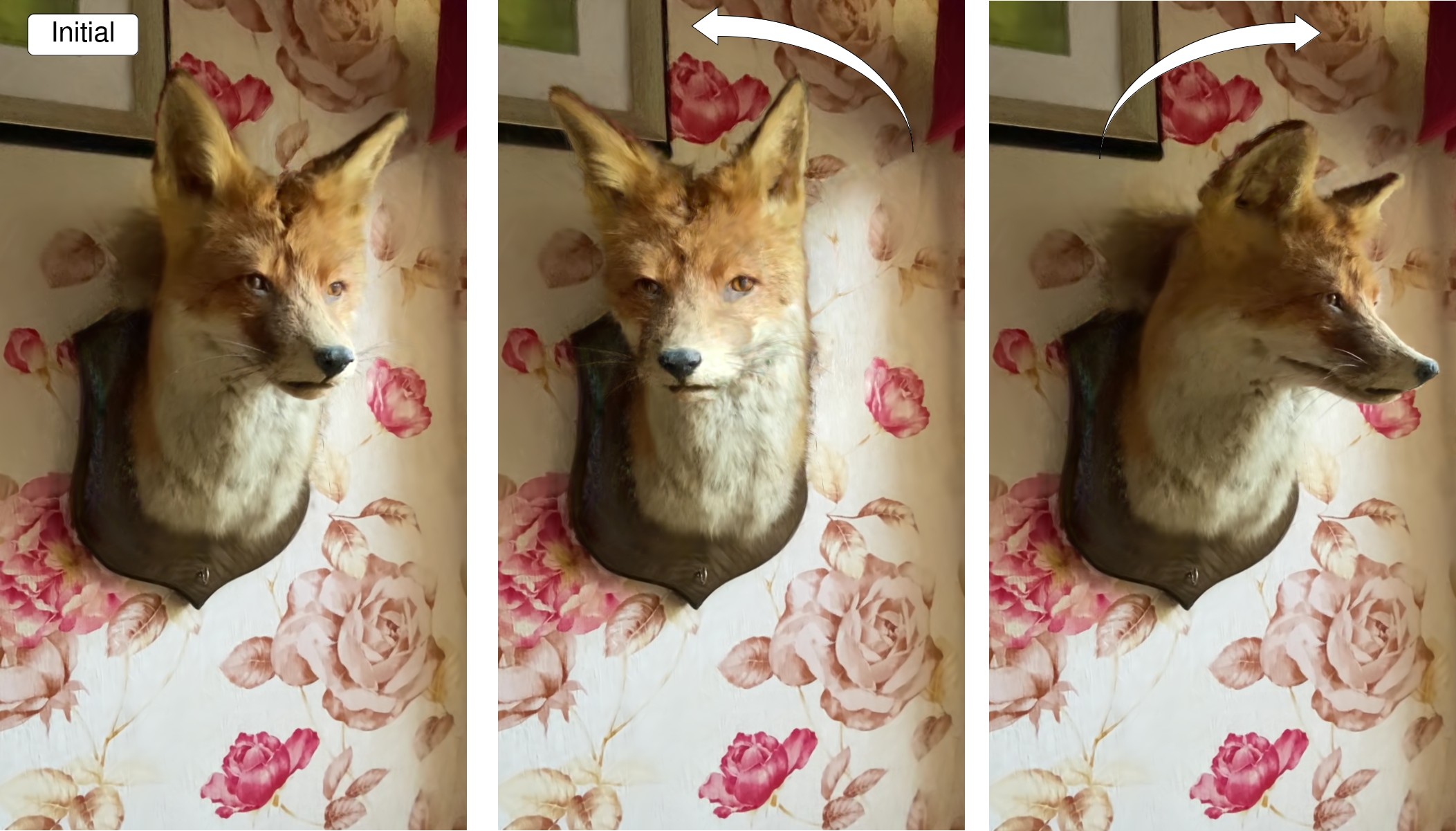}
    \caption{Blender provides a wide range of physical properties and force sources, including gravity and wind, to create realistic and dynamic animations. In this experiment, we present examples of physics animation in object scenes, showing the versatility of animating both small objects, such as a flower on a table, and large objects, such as a fox's head dominating the scene.} 
    
\label{fig:kwiatek} 
% \end{figure}
\end{wrapfigure}

Achieving high-quality renders with Gaussian Splatting methods typically requires a substantial number of Gaussians. Previously, animations were executed directly on Gaussians \cite{xie2024physgaussian} or using estimated meshes \cite{guedon2024sugar}. However, these simulations demand significant time and computational resources. In the Genesis engine \cite{Genesis}, this process is optimized by employing simulations facilitated through particles selected via Sampler component, lowering the number of points. To achieve a similar mechanism, we employ BIRCH clustering to construct Gaussians called Core-Gaussians for each cluster.
%on Gaussians, solvsimulationsers
This enables the creation of a hierarchical structure for Gaussians. The simulation is then performed only on Core Gaussians, and then their move pattern is transferred on Sub Gaussians inspired by \cite{waczynska2024d}, see Fig. \ref{fig:hier}. We demonstrate that physics engines can be applied to selected Gaussians without compromising visual quality (see Fig. \ref{fig:hier}). This approach significantly accelerates the simulation process while maintaining rendering quality.

%%%%%%%%%%%%%%%%%%%%%%%%%%%%5
%%%%%%%%%%%%%%%%%%%%%%%%
%%%%%%%%%%%%%%%%%%%%%%%%%%%%%%
%%%%%%%%%%%%%%%%%%%%%%%%%%%%%%
%%%%%%%%%%%%%%%%%%%%%%%%%%%%%%%%
\paragraph{\our{} for Dynamic Scene}
\our{} is a natural combination of a physics engine dedicated to points and Gaussian Splatting with flat Gaussians. The reparametrization in GaMeS can be applied to a multitude of disparate methodologies, with a particular focus on flat Gaussian methods. Moreover, it can be employed for dynamic Gaussian Splatting. In \cite{waczynska2024d}, the D-MiSo model is introduced, which uses flat Gaussians for the modeling of dynamic scenes. Such representations can be utilized in conjunction with the aforementioned approach. In the D-MiSo framework, two distinct types of Gaussians and two deformable networks are employed. In the process of inference, only sub-Gaussians are employed, which are modified by deformable function composition $\psi(v,s)$, i.e.,
\begin{equation}
    \G_{\text{D-MiSo}} =  \left\{ \left( \N_{sub} \left( \psi (V_j^i,s) \right), \sigma_i, c_i  \right) \right\}_{i=1}^{p},
\end{equation}
where $s$ represents a time value derived from the dynamics observed in the scenes. In our solution, we can incorporate a deformation map $\phi(X,t)$, which comes from the physics simulator. Consequently, \our{} for the dynamic scene is given as follows:
\begin{equation}
    \G_{t} =  \left\{ \left( \N_{sub} \left( \phi( \psi (V_j^i,s),t) \right), \sigma_i, c_i  \right) \right\}_{i=1}^{p},
\end{equation}
where $s$ represents time in the dynamic scene and $t$ denotes time in the simulation.
%\text{\our{}-D-MiSo}

\section{Experiments}
% The experiments section shows that \our{} is a universal method, which works with small objects (Fig.~\ref{fig:mis}) and large scenes (Fig.~\ref{fig:kwiatek}). Moreover, we can model the interaction between objects (see Fig.~\ref{fig:miswkubku}) as well as break objects into pieces (see Fig.~\ref{fig:kubek}). \our{} work also with dynamic scenes Fig.~\ref{fig:standup}. 

This experimental section demonstrates that \our{} is a universal method that can be applied to small objects and large scenes alike. Furthermore, it allows for the modeling of interactions between objects and breaking them into pieces. \our{} is also capable of working with dynamic scenes. 

% The experiments section is divided into three parts. Firs we show how our model work on synthetic data. Then we show physics on large scenes. Finally we show \our{} in the case of dynamic scenes. 

% \subsection{Implementation Details}
% Here, we provide a detailed overview of our implementation and a comprehensive description of the datasets used as benchmarks during our experiments. The corresponding source code is publicly accessible on GitHub {\color{red}missing link}. Our codebase was developed starting from the GS \cite{kerbl20233d} vanilla code, adhering strictly to its licensing terms. The computational experiments were executed on NVIDIA GeForce RTX 4090 and A100 GPUs, and primarily focused on simulations conducted using Taichi \cite{Taichi} or Blender with its lattice deformation tool. %\asia{dodac info o latice(?) nie pamietam jak to sie nazywa}.
% {\color{red}Warto napisac co to Taichi i dlaczego Blender.}

% \subsection{Datasets}
% We conducted our experiments to highlight the fundamental advantages of our model, employing a variety of well-established datasets, including Synthetic Dataset, as well as more complex large-scale datasets such as the Mip-NeRF360. Additionally, we illustrated the strengths of our approach further with several specific objects. These objects are distributed under a CC-BY or CC-0 license from Blend Swap\footnote{https://blendswap.com/}.

\begin{figure}[t]
  \centering
  %\begin{minipage}[t]{0.45\textwidth}
    \begin{tabular}[h]{ccc}
    $t_0$  &  $t_i$ & $t_T$ \\
    \includegraphics[width=0.25\linewidth]{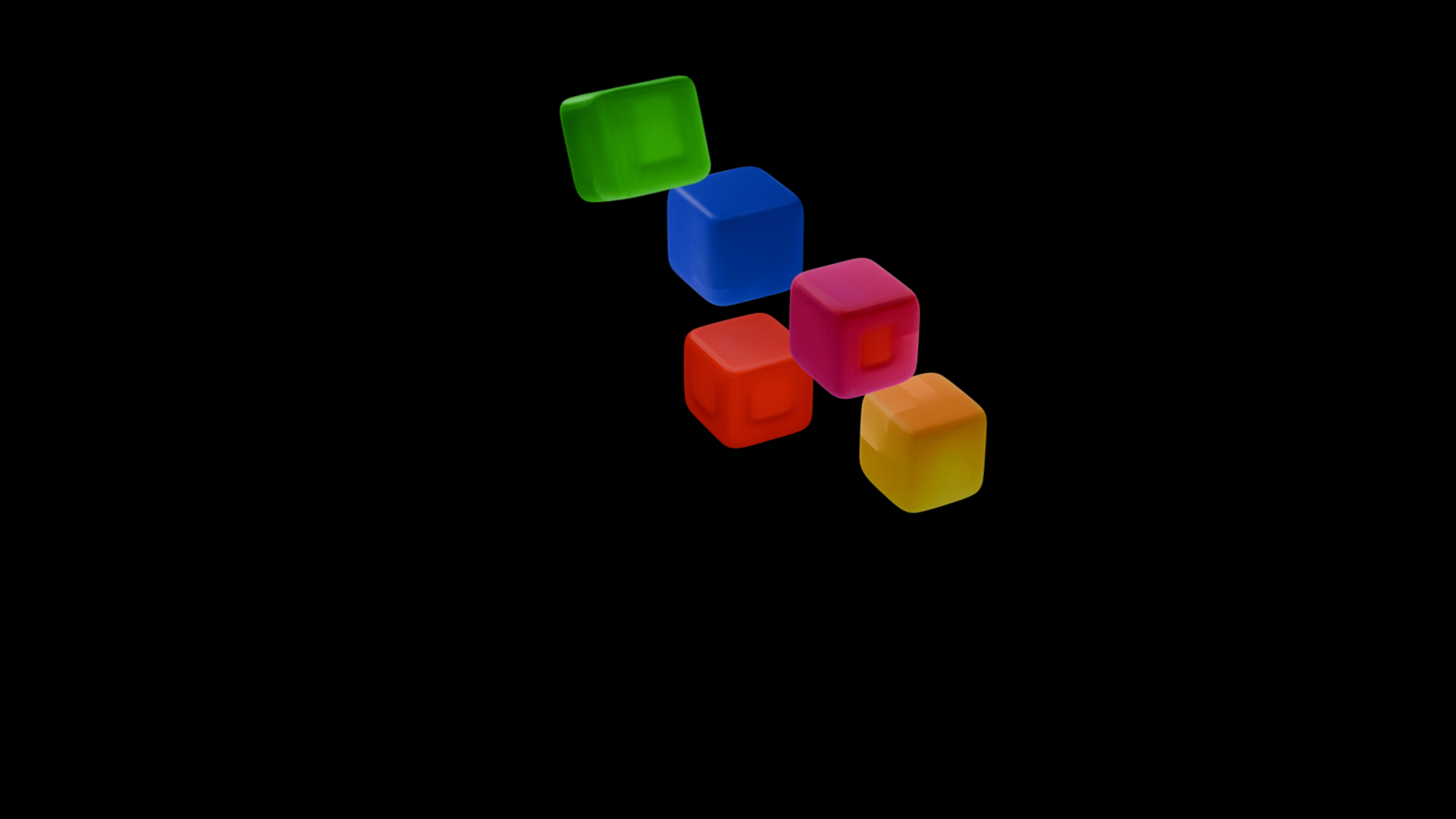} &
    \includegraphics[width=0.25\linewidth]{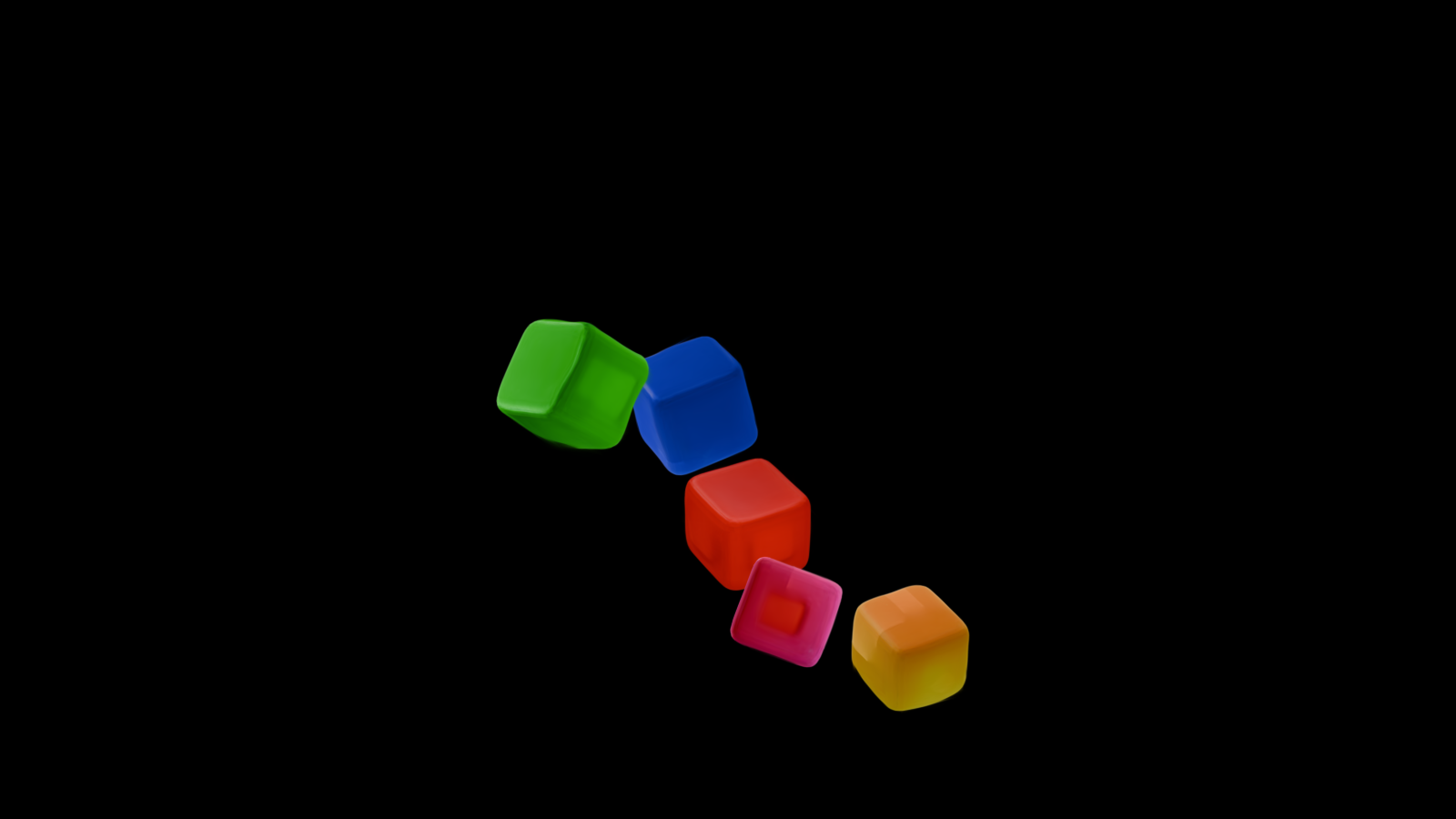} &
    \includegraphics[width=0.25\linewidth]{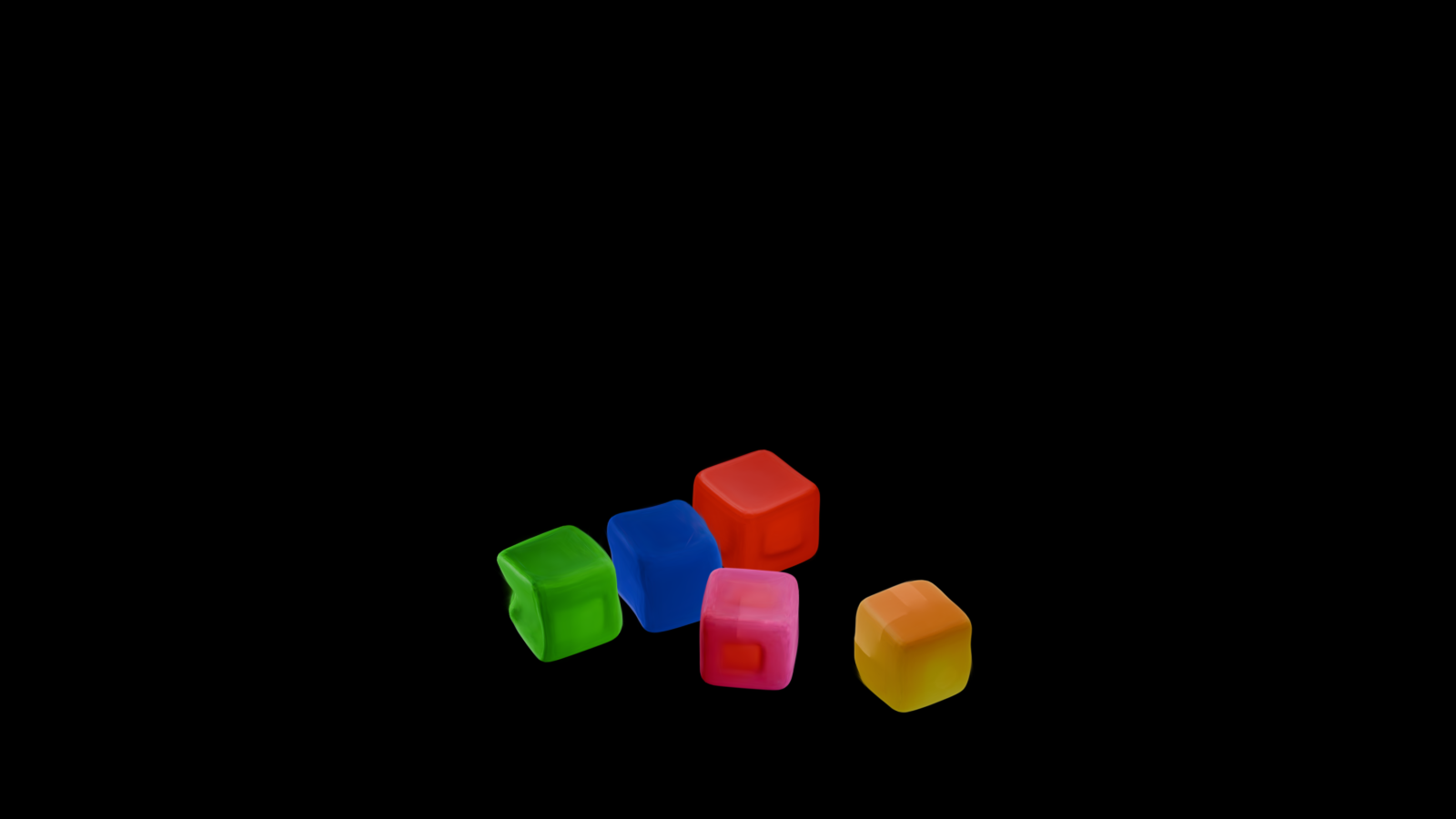}
\\
    \includegraphics[width=0.25\linewidth]{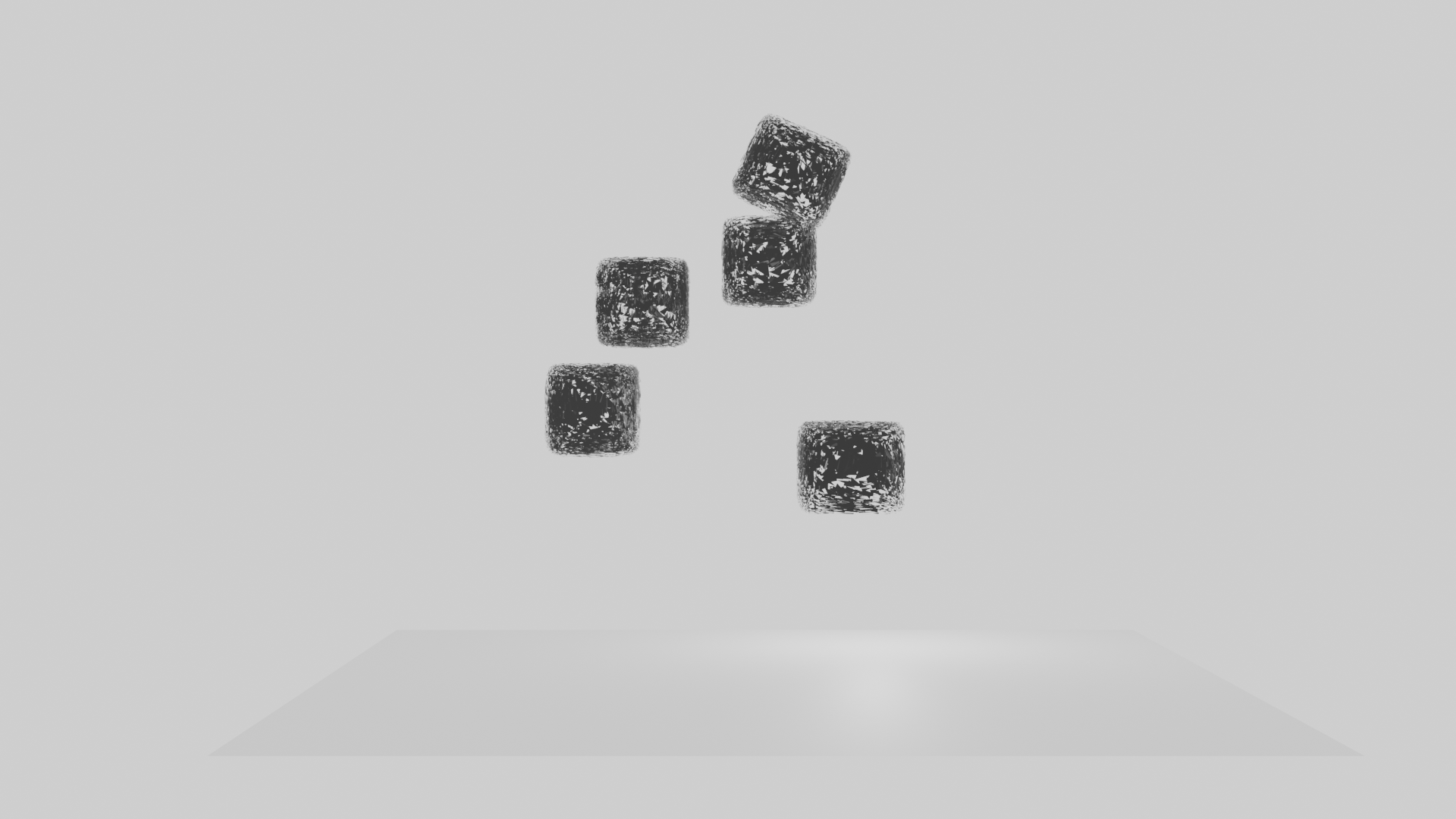}&
    \includegraphics[width=0.25\linewidth]{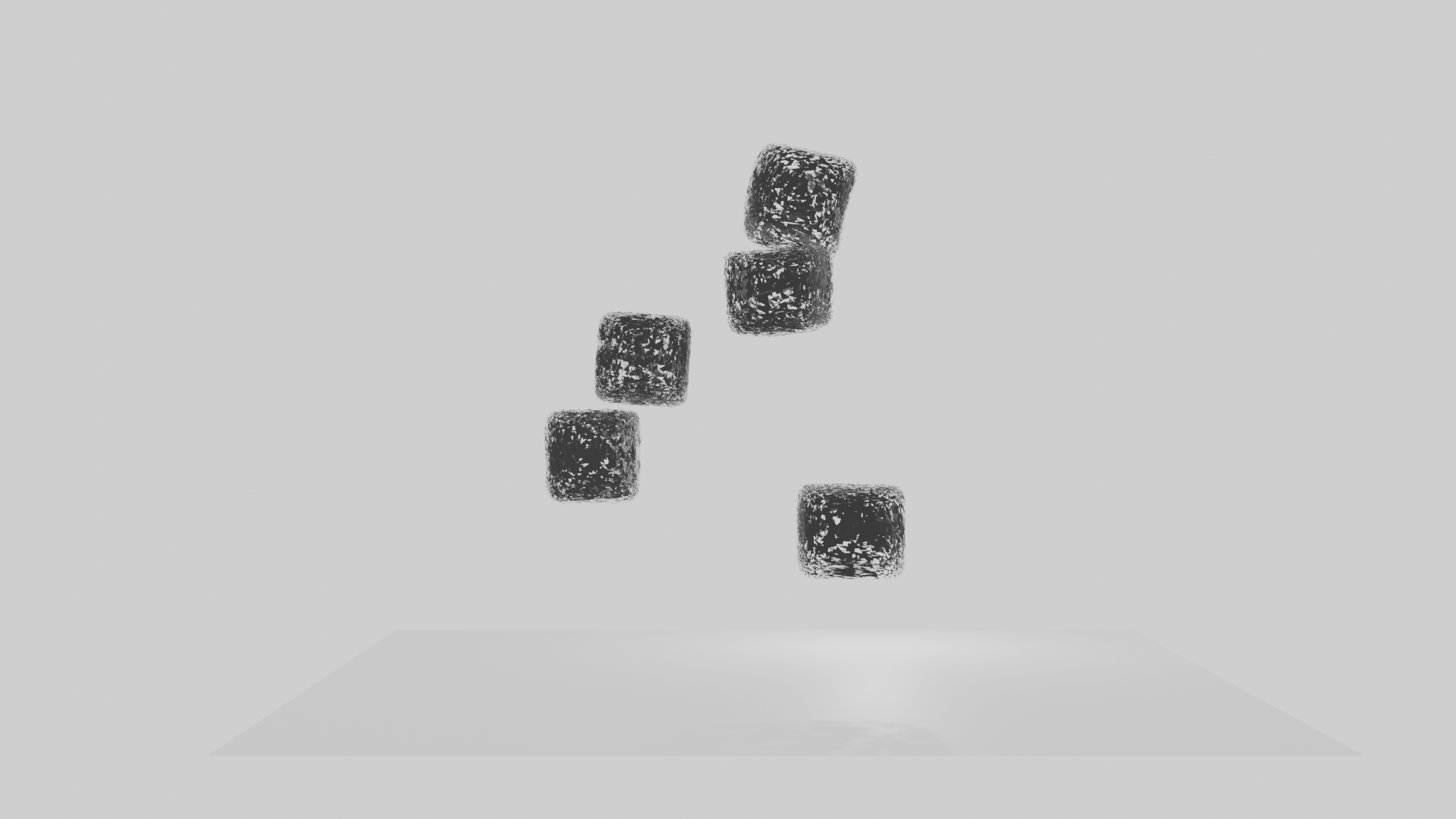}&
    \includegraphics[width=0.25\linewidth]{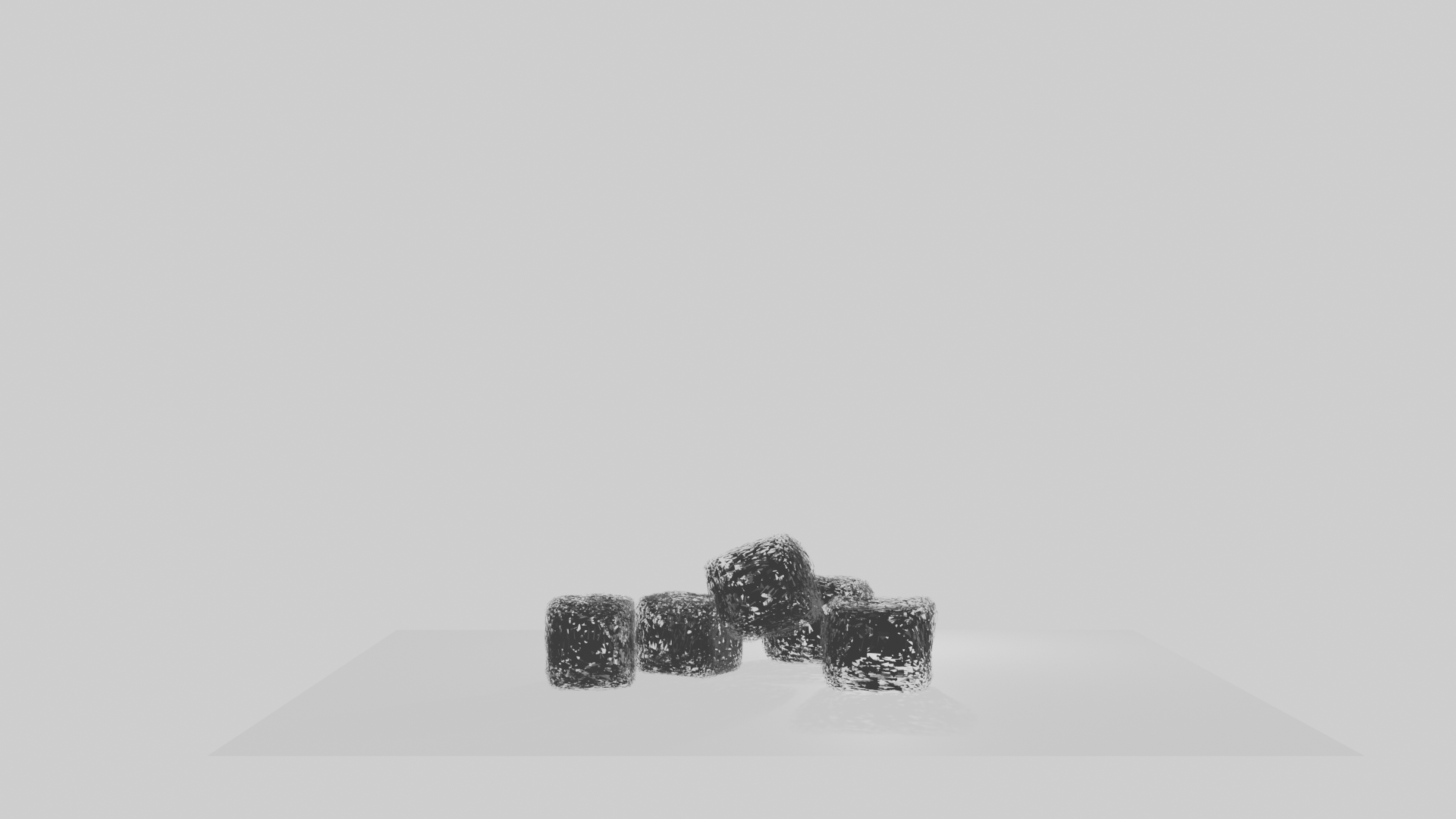}

\end{tabular}
    \caption{\new{Jelly cube falling simulation created in Blender using soft-body and cloth modifiers to mimic gel-like behavior. In this setup, each cube was trained independently with GaMeS, allowing every cube in the scene to exhibit distinct properties and more realistic collision interactions.}}
    \label{fig:cubes}
  %\end{minipage}
  %\hfill
  %\begin{minipage}[t]{0.45\textwidth}
\end{figure}

\begin{figure}[t]
  \centering
  %\begin{minipage}[t]{0.45\textwidth}
    \begin{tabular}[h]{ccc}
    $t_0$  &  $t_i$ & $t_T$ \\
    \includegraphics[width=0.25\linewidth]{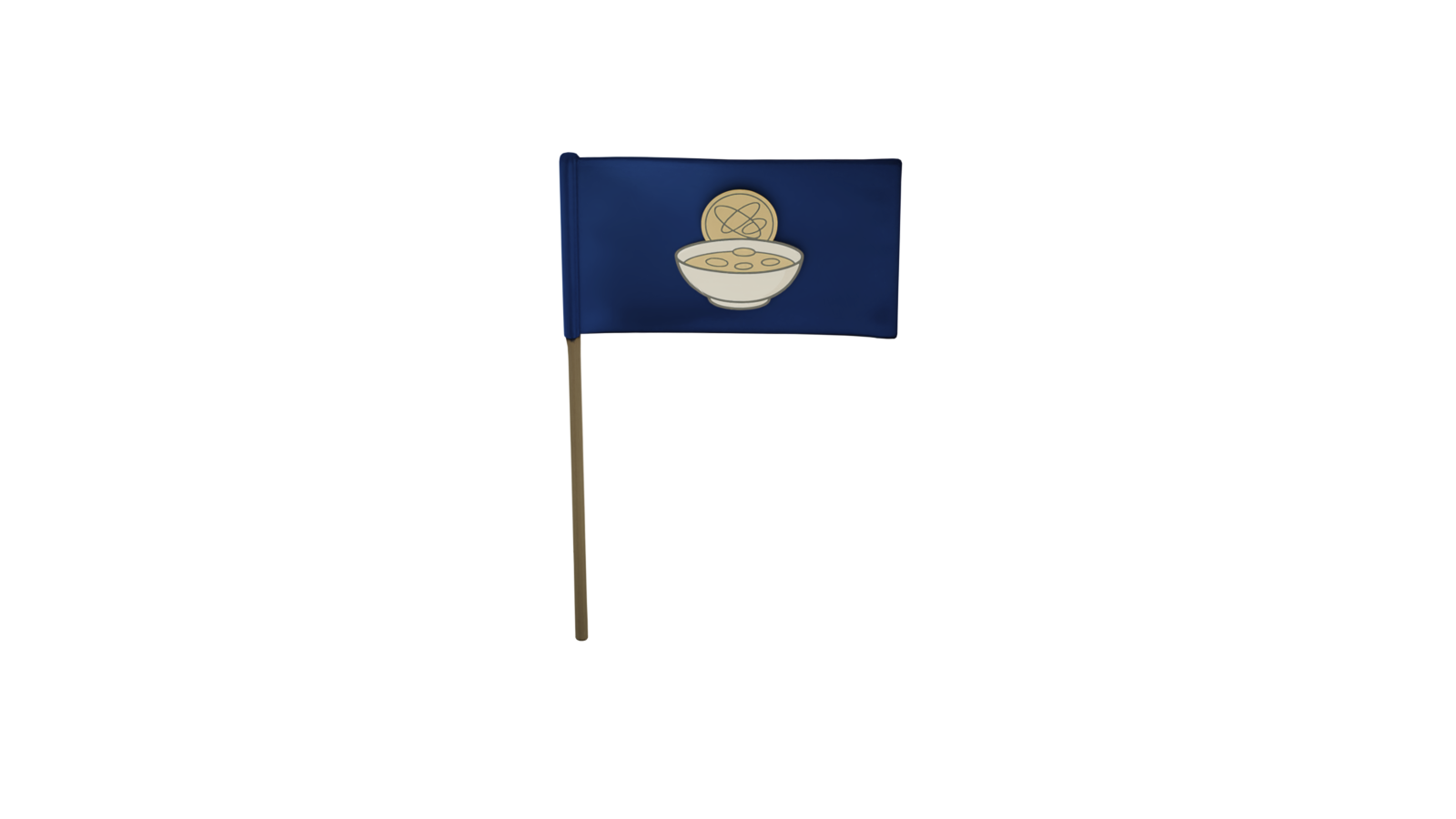} &
    \includegraphics[width=0.25\linewidth]{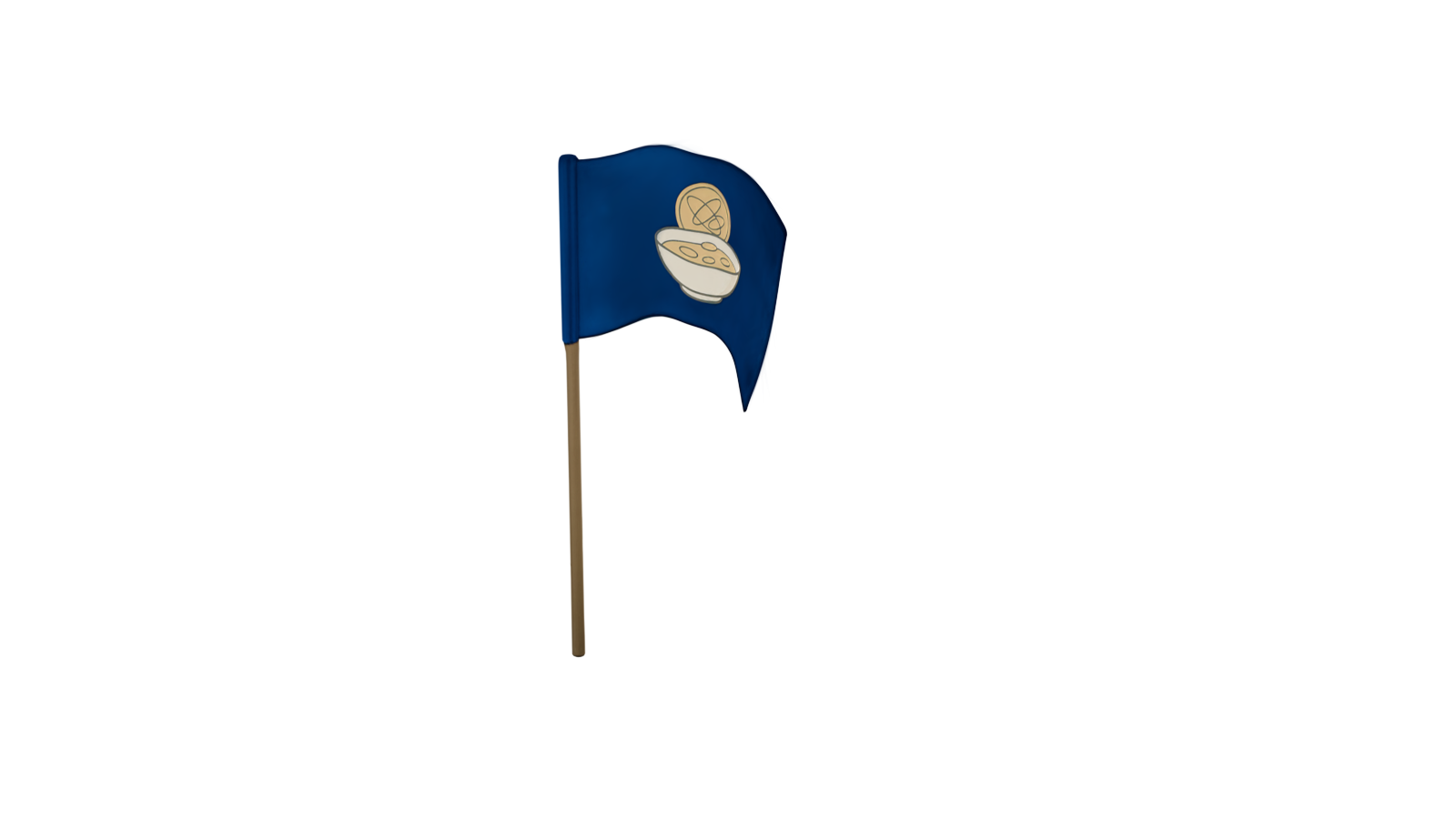} &
    \includegraphics[width=0.25\linewidth]{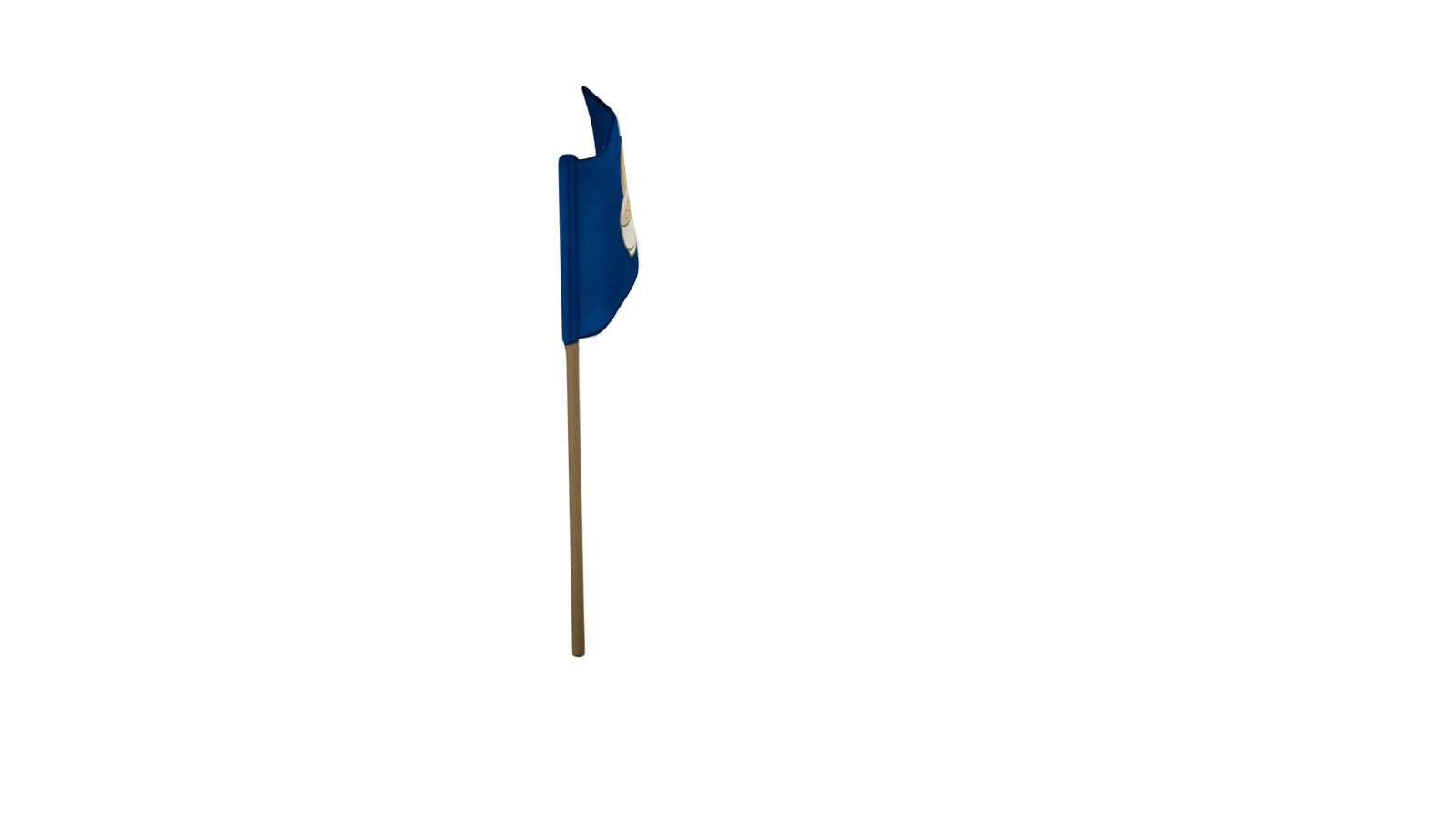}
\\
    \includegraphics[width=0.25\linewidth]{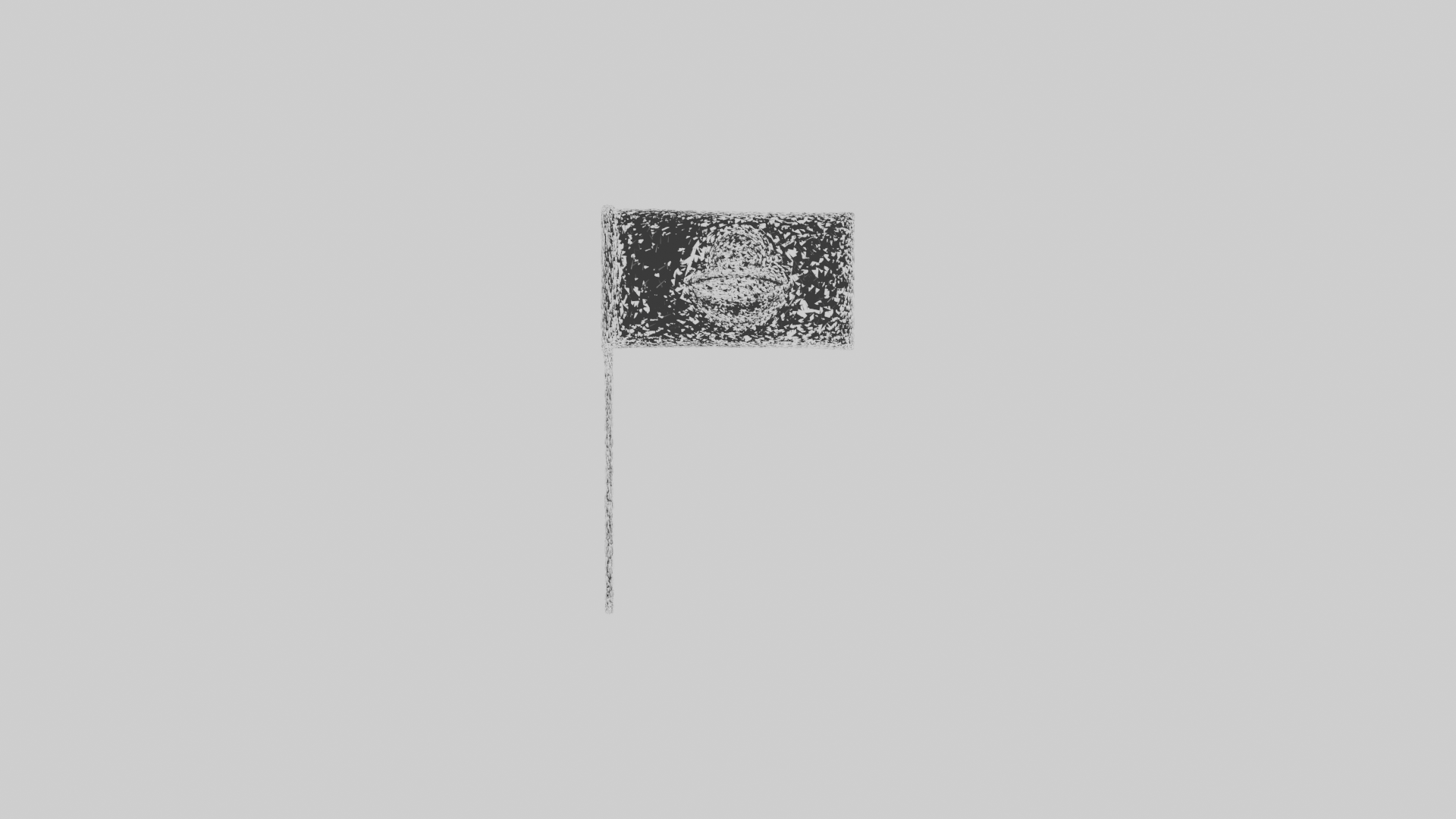} &
    \includegraphics[width=0.25\linewidth]{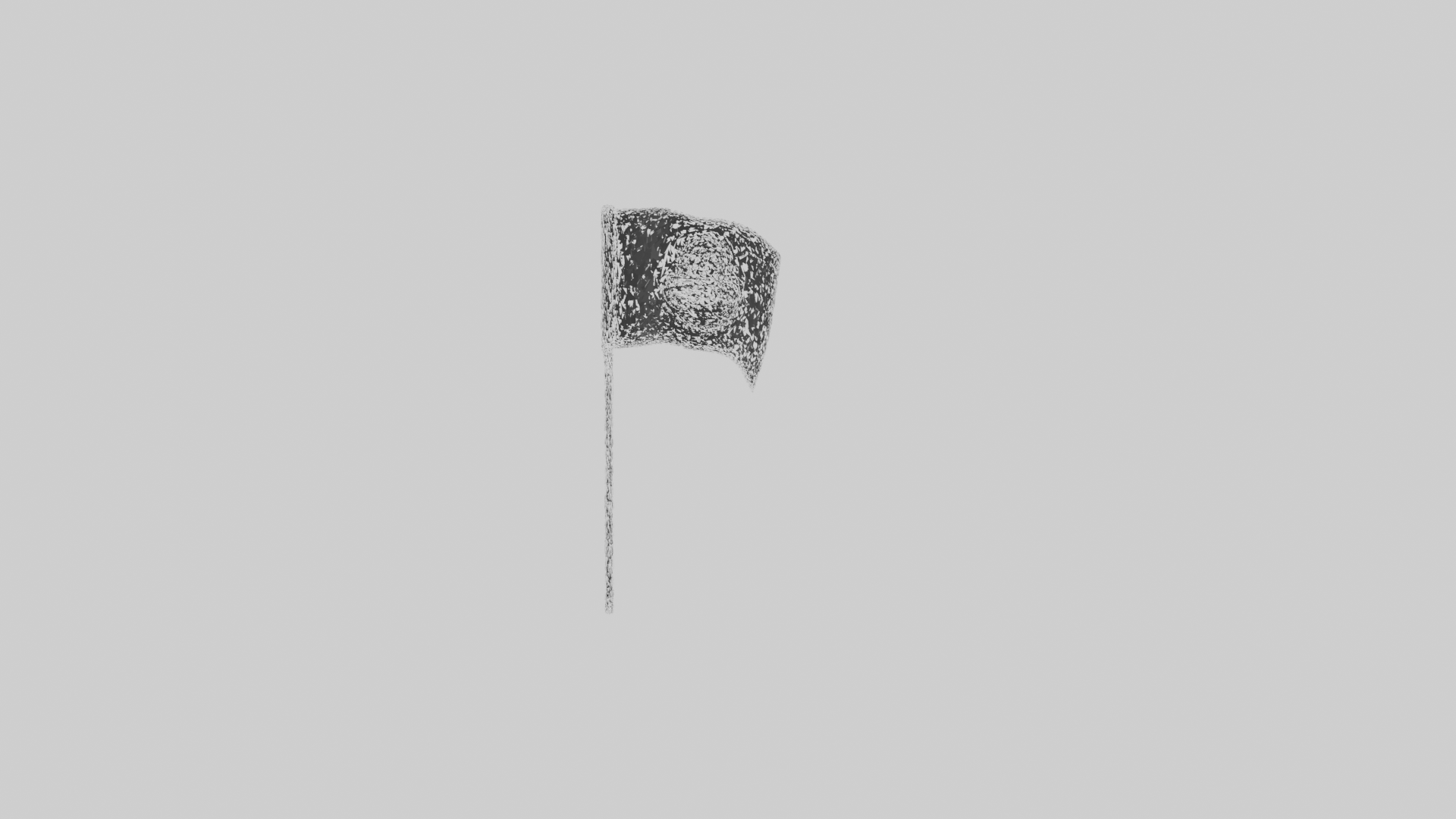} &
    \includegraphics[width=0.25\linewidth]{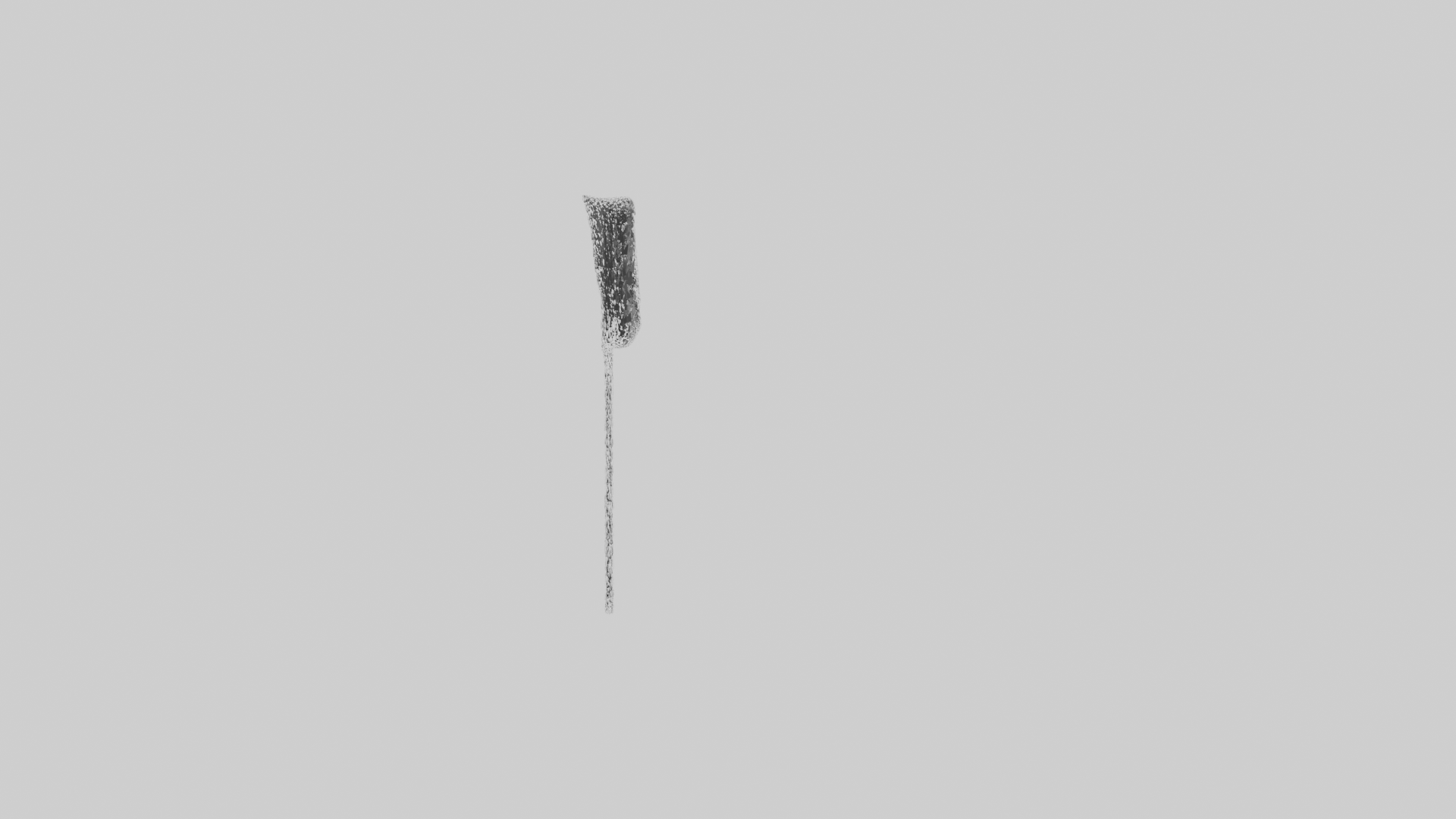}

\end{tabular}
    \caption{\new{Flag movement simulation created in Blender using the Lattice modifier, based on a triangle soup.}}
    \label{fig:flag}
  %\end{minipage}
  %\hfill
  %\begin{minipage}[t]{0.45\textwidth}
\end{figure}

\paragraph{Static Objects}

The Shiny Dataset~\cite{verbin2022refnerf} contains examples of static objects with varying textures. In such a case, all Gaussian components are converted to GaMeS-based representations, after which the physics engine is used to modify their positions. Specifically, \our{} is employed to move these representations in accordance with a dynamic function. As can be observed in Fig.~\ref{fig:tesser}, Fig.~\ref{fig:kaczki}, Fig.~\ref{fig:kwiatek}, Fig.~\ref{fig:cubes}, and Fig.~\ref{fig:flag} the model can produce high-quality animation renders. 

\begin{figure}[t]
    \centering
       \includegraphics[width=0.65\textwidth, trim=0 0 100 0, clip]{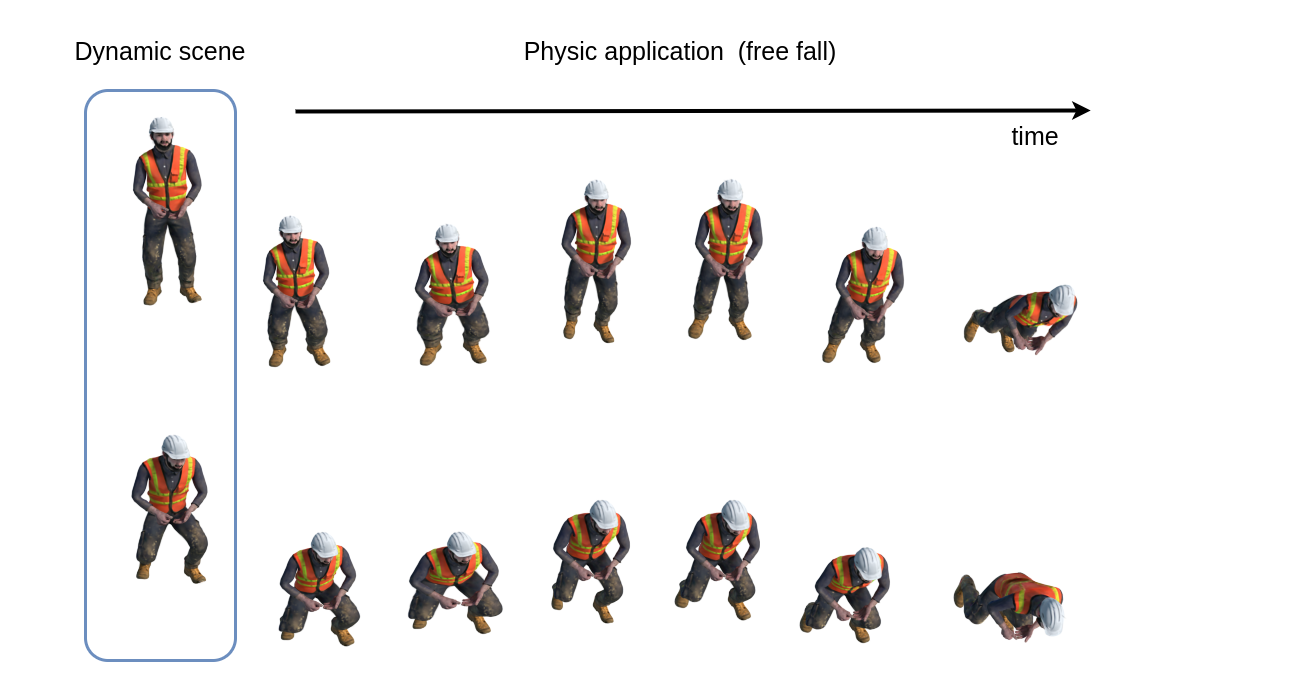}
    \caption{
    Controlling animations using physics engines allows for application to dynamic objects. This figure compares animations of a dynamic object at two different times under the same physical parameters. This experiment uses \our{} integrated with the D-MiSo model~\cite{waczynska2024d}.} 
\label{fig:standup} 
%\vspace{-0.5cm}
\end{figure}

\paragraph{Dynamic Objects} 
We show that our pipeline also works with dynamic scenes using sample objects from the D-NeRF dataset~\cite{pumarola2020d}. We use D-MiSo~\cite{waczynska2024d} to model dynamic scenes using flat Gaussians. Then, we select a frame and use a physics engine. Fig.~\ref{fig:tesser} and Fig.~\ref{fig:standup} present how the objects change according to the physics engine and timeline. %Additionally, in Tab.~\ref{tab_2} we show the FID score in three different time steps of dynamic scenes, which were obtained by \our{}. As we can see, the renders have similar scores, which means that our model produces consistent simulations.

\paragraph{Multiple Objects}
We show that \our{} can model the interaction of static objects, as shown in Fig. \ref{fig:gaussmoves}. In this case, each element of the scene is trained separately. In addition, the properties of each component are specified by different parameters. Fig.~\ref{fig:tesser} shows the application of physics to two interacting objects: a falling ball and a car reacting to its impact. In Fig.~\ref{fig:kaczki}, we integrate small duck objects into a full scene. As can be seen, the ducks bounce off the table and fall to the ground. In Fig.~\ref{fig:artifacts}, we show a sand-based teddy bear and a glass cup. As we can see, the sand falls smoothly into the cup. In the same figure, we also show how our model handles the formation of artifacts.

\paragraph{Model of Physics}
We demonstrate that physics modeling can function as a black box by using three different simulation approaches.

Genesis\footnote{https://github.com/Genesis-Embodied-AI/Genesis} \cite{Genesis} is a versatile physics simulation platform designed for Robotics, Embodied AI, and Physical AI applications, featuring a robust physics engine, ultra-fast robotics simulation. It supports diverse physics solvers including, but not limited to, MPM, FEM, PBD, as well as various material models. In our work, we focus on the MPM solver.

Taichi$\_$elements\footnote{https://github.com/taichi-dev/taichi$\_$elements} is a highly efficient physics engine for the simulation of multi-material continuum mechanics, originally developed in Python using taichi \cite{hu2019taichi}, which is designed for high-performance parallel numerical computation. It supports multiple materials, including water, elastic objects, snow, and sand.

\begin{figure}[t]
    \centering
    \includegraphics[width=0.6\textwidth, trim=0 0 0 0, clip]{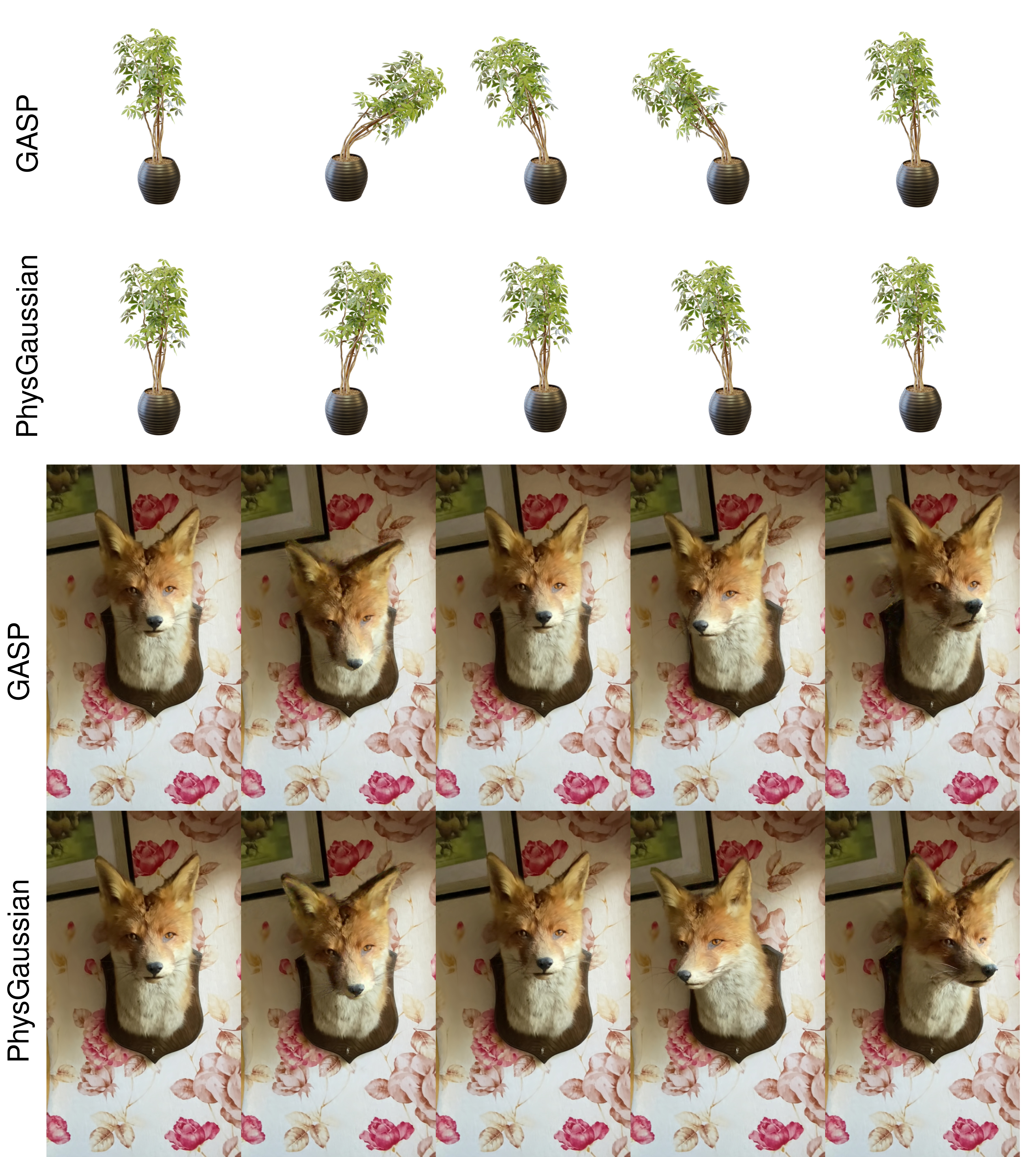}
    \caption{Comparison of renders obtained by \our{} and PhysGaussian on ficus and fox simulations. For both methods, the same trained model was used, and the simulations were conducted using the physics engine used in PhysGaussian. For \our{} we used vertices of triangle soup as positions used during simulation, after each step their position was used to calculate the parameters of Gaussians represented by each triangle. } 
   
\label{fig:physgaussan} 
\end{figure}

We also carried out an experiment to visually and numerically compare our method with PhysGaussian (see Fig. \ref{fig:physgaussan}). Specifically, we trained a GaMeS model and integrated it directly into their engine. Their approach performs MPM simulations on Gaussian centers followed by covariance calculations. In contrast, we utilized their engine but performed simulations on designated points from Gaussians without additional covariance computations. In Tab. \ref{tab_1} shows that, qualitatively, the FID remains consistent with PhysGaussian. Since we use the same engine and the same number of Gaussians, the result is expected and maintains fairness in comparison. Tab. \ref{time} shows that incorporating hierarchical structuring significantly accelerates the simulation process. Moreover, visual comparison is included in Fig.~\ref{fig:physgaussan}.

\begin{table}[t]
    \centering
    % \scriptsize
%    \fontsize{7.5pt}{11pt}\selectfont
    %\caption{Fréchet Inception Distance (FID, lower is better) calculated in static scenes (Vasedeck, Fox, Teddybear, Ficus, Bottle).}
    %\label{tbl:fid}
    %\begin{subtable}[t]{0.4\textwidth}
    \centering
        \begin{tabular}{cccccc}
        \toprule
          % Method & Bonsai & Fox \\
          % \midrule
          % GS & 19.71 & 51.04 \\
          % GaMeS & 19.64 & 47.62\\
          % \our{} (our) & 38.33 & 88.59\\
        Method & Vasedeck & Fox & Teddybear & Ficus & Bottle\\
          \midrule
          PhysGaussian & 68.66 & 167.16 & 227.91 & 86.63 & 261.48 \\
          \our{} (our) & 68.22 & 168.55 & 244.39 & 83.98 & 254.84 \\         
        \bottomrule
        \end{tabular}
               \caption{FID score obtained by PhysGaussian~\cite{xie2024physgaussian} and \our{} for static scenes and renders on images taken across simulations. For both methods, we use the same physics engine and trained GS model. In the case of \our{}, we treated each triangle vertex as a Gaussian and used their positions across the simulation to calculate the actual parameters of each Gaussian.}
        \label{tab_1}
    %\end{subtable} 
    %\qquad
    %\begin{subtable}[t]{0.2\textwidth}
    %\centering
     %   \begin{tabular}{cc}
     %   \toprule
     %     Frame & Stand-up \\
     %     \midrule
     %     Start & 133.55  \\
     %     Mid   & 112.79 \\
     %     End   & 142.90 \\
     %     \bottomrule
     %   \end{tabular}
     %   \caption{FID score for three different time steps obtained by \our{}. Note that the simulation is consistent across dynamic scenes.}
      %  \label{tab_2}
    %\end{subtable}
    %\label{tab}
    % \vspace{-0.2cm}
\end{table}

\begin{table}[t]
    \centering
    % \scriptsize
    \fontsize{7.5pt}{11pt}\selectfont
    %\caption{Fréchet Inception Distance (FID, lower is better) calculated in static scenes (Vasedeck, Fox, Teddybear, Ficus, Bottle).}
    %\label{tbl:fid}
    %\begin{subtable}[t]{0.4\textwidth}
    \centering
        \begin{tabular}{cccccc}
        \toprule
          % Method & Bonsai & Fox \\
          % \midrule
          % GS & 19.71 & 51.04 \\
          % GaMeS & 19.64 & 47.62\\
          % \our{} (our) & 38.33 & 88.59\\
        Method & Vasedeck & Fox & Teddybear & Ficus & Bottle\\
          \midrule
          PhysGaussian & 53.82 & 110.37 & 618.07 & 89.29 & 46.42\\ 
          \our{} (our) & 74.55 & 168.81 & 614.55 & 186.44 & 87.03\\  
           \our{} (our with Core) & 17.94 & 64.69 & 351.18 & 19.93 & 18.80
           \\
        \bottomrule
        \end{tabular}
        \caption{Comparison of time in seconds required to finish the simulation for various models between PhysGaussian and \our{}. Both methods use the same GS models and physics engine. For \our{}, we also show the speed for simulations where only Core Gaussians were used during simulation.}
        \label{time}
    %\end{subtable} 
    %\qquad
    %\begin{subtable}[t]{0.2\textwidth}
    %\centering
     %   \begin{tabular}{cc}
     %   \toprule
     %     Frame & Stand-up \\
     %     \midrule
     %     Start & 133.55  \\
     %     Mid   & 112.79 \\
     %     End   & 142.90 \\
     %     \bottomrule
     %   \end{tabular}
     %   \caption{FID score for three different time steps obtained by \our{}. Note that the simulation is consistent across dynamic scenes.}
      %  \label{tab_2}
    %\end{subtable}
    %\label{tab}
    % \vspace{-0.2cm}
\end{table}

\begin{table}[b]
    \centering
    \begin{tabular}{ccccccc}
    \toprule
    $\alpha$ & 1.0 & 2.0 & 3.0 & 4.0 & 5.0 & $\infty$ \\
    \midrule
    FID & 156.99 & 162.91 & 168.21 & 168.73 & 168.59 & 188.21 \\
    \bottomrule
    \end{tabular}
    \caption{\new{Ablation study of the proposed strategy with varying $\alpha$ values. The FID score generally increases with larger $\alpha$, highlighting the effectiveness of the method. $\infty$ indicates the strategy was not applied.
}
    }
    \label{tab:ablation_alpha}
\end{table}

\new{We conduct an ablation study to assess the impact of the proposed Gaussian Control Rule. The results, summarized in Table~\ref{tab:ablation_alpha}, report the value of $\alpha$ used in Equation~\ref{eq:alpha_rule} (first row) and the corresponding FID score on the simulation shown in Figure~\ref{fig:artifacts} (second row). The findings indicate that the strategy yields consistent improvements. Furthermore, the FID score generally increases with larger $\alpha$ values, with only a slight drop observed between $\alpha=4.0$ and $\alpha=5.0$.
}

\section{Conclusion}
In this paper, we introduce \our{}, a pipeline that integrates high-quality rendering with accurate physics simulation. Using a GaMeS-based representation of objects, our approach enables the simulation of basic phenomena, such as crushing, and more complex scenarios, including cracking. This demonstrates the model's versatility and effectiveness in handling a wide range of physics interactions within simulated environments. Conducted experiments show that \our{} works in many different scenarios and produces high-quality simulations. We observe that the additional control rules we proposed for hierarchical simulations significantly accelerate the simulation process. This approach resembles the particle sampler commonly utilized in many physics engines.

\paragraph{Limitations} 

\new{Our model may produce artifacts when the simulation causes drastic changes in the shape of the original object. Moreover, the \our{} pipeline does not account for variations in lighting interaction with the modified surface. Since our method is built upon the GaMeS representation, it also inherits its limitations. Flat Gaussians may require a large number of components to accurately capture highly curved or anisotropic regions, while the mesh-free triangle soup lacks explicit connectivity. On the other hand, in mesh-based settings, the representation quality is constrained by the resolution of the mesh.}

\section{Acknowledgements}
The project ``Effective rendering of 3D objects using Gaussian Splatting in an Augmented Reality environment'' (FENG.02.02-IP.05-0114/23) is carried out within the First Team programme of the Foundation for Polish Science co-financed by the European Union under the European Funds for Smart Economy 2021-2027 (FENG).
The work of P. Spurek was supported by the National Science Centre (Poland), Grant No. 2021/43/B/ST6/01456. 

% Since our presented pipeline depends on the learned GS-based model, and an object is not well-represented in the training set, the model may fail to generate an accurate spatial representation. This inherited from GS limitation impacts the physical simulation, as it lacks access to a precise visualization of the object or scene, leading to potential inaccuracies in the simulation results.

%\newpage

%{\small
%\bibliographystyle{ieee_fullname}
%\bibliography{ref}
%}

\end{document}